\documentclass[journal]{IEEEtran}

\ifCLASSINFOpdf
\else
\fi

\usepackage[compact]{titlesec}
\usepackage{subfig}
\usepackage{graphicx} 
\usepackage[noend]{algorithm2e}
\usepackage{algorithmic}
\usepackage{setspace}
\usepackage{url}
\usepackage{amsmath,amsthm}
\usepackage{amssymb,wasysym}
\usepackage{mathrsfs}
\usepackage{cite}
\usepackage{xcolor}
\usepackage{wrapfig}
\usepackage{verbatim}
\usepackage{dsfont}
\usepackage{multirow}
\usepackage{breqn,xspace}

\usepackage{amsmath} 
\usepackage{amssymb} 
\usepackage{amsfonts} 

\usepackage{multirow}
\usepackage{booktabs}
\usepackage{graphicx}
\usepackage[compact]{titlesec}

\ifodd 1

\else

\fi

\newcommand{\name}{HSplitLoRA\xspace}

\begin{document}

\title{\name: A Heterogeneous Split Parameter-Efficient Fine-Tuning Framework for Large Language Models}

\author{Zheng Lin, Yuxin Zhang, Zhe Chen,~\IEEEmembership{Member,~IEEE}, Zihan Fang,  Xianhao Chen,~\IEEEmembership{Member,~IEEE}, Praneeth Vepakomma,~\IEEEmembership{Member,~IEEE},  Wei Ni,~\IEEEmembership{Fellow,~IEEE}, Jun Luo,~\IEEEmembership{Fellow,~IEEE}, and Yue Gao,~\IEEEmembership{Fellow,~IEEE}

\thanks{Z. Lin, Y. Zhang, Z. Chen, Z. Fang, and Y. Gao are with the Institute of Space Internet, Fudan University, Shanghai 200438, China, and the School of Computer Science, Fudan University, Shanghai 200438, China (e-mail: zlin20@fudan.edu.cn; zhfang19@fudan.edu.cn; zhechen@fudan.edu.cn; yxzhang24@m.fudan.edu.cn;  gao.yue@fudan.edu.cn). Z. Lin is also with the Department of Electrical and Electronic Engineering, University of Hong Kong, Pok Fu Lam, Hong Kong, China.}
\thanks{X. Chen is with the Department of Electrical and Electronic Engineering, University of Hong Kong, Pok Fu Lam, Hong Kong, China (e-mail:
xchen@eee.hku.hk).}
\thanks{P. Vepakomma is with Mohamed bin Zayed University of Artificial Intelligence, Abu Dhabi, United Arab Emirates, and the Massachusetts Institute of Technology, Cambridge, MA 02139 USA (e-mail: vepakom@mit.edu).}
\thanks{W. Ni is with Data61, CSIRO, Marsfield, NSW 2122, Australia, and the School of Computing Science and Engineering, and the University of New South Wales, Kennington, NSW 2052, Australia (e-mail:
wei.ni@ieee.org).}
\thanks{J. Luo is with the School of Computer Engineering, Nanyang Technological University, Singapore (e-mail: junluo@ntu.edu.sg).}
\thanks{\textit{(Corresponding author: Yue Gao)}}
}



\maketitle

\begin{abstract}
Recently, large language models (LLMs) have achieved remarkable breakthroughs, revolutionizing the natural language processing domain and beyond. Due to immense parameter sizes, fine-tuning these models with private data for diverse downstream tasks has become mainstream. Though federated learning (FL) offers a promising solution for fine-tuning LLMs without sharing raw data, substantial computing costs hinder its democratization. Moreover, in real-world scenarios, private client devices often possess heterogeneous computing resources, further complicating LLM fine-tuning. To combat these challenges, we propose \name, a heterogeneous parameter-efficient fine-tuning (PEFT) framework built on split learning (SL) and low-rank adaptation (LoRA) fine-tuning, for efficiently fine-tuning LLMs on heterogeneous client devices. \name first identifies important weights based on their contributions to LLM training. It then dynamically configures the decomposition ranks of LoRA adapters for selected weights and determines the model split point according to varying computing budgets of client devices. Finally, a noise-free adapter aggregation mechanism is devised to support heterogeneous adapter aggregation without introducing noise. Extensive experiments demonstrate that \name outperforms state-of-the-art benchmarks in training accuracy and convergence speed. 
\end{abstract}

\begin{IEEEkeywords}
Distributed learning, split learning, large language model, parameter-efficient fine-tuning.
\end{IEEEkeywords}

\IEEEpeerreviewmaketitle

\section{Introduction}

Recently, large language models (LLMs) have achieved tremendous success across a broad spectrum of pivotal sectors due to their exceptional ability in handling high-complexity and large-scale datasets~\cite{achiam2023gpt,team2023gemini,touvron2023llama,lin2024splitlora,fang2024automated}. Notable examples include OpenAI's  GPT series~\cite{achiam2023gpt}, Google's Gemini~\cite{team2023gemini}, and Meta's LLaMA~\cite{touvron2023llama}, all of which have gained significant attention in both industry and academia. These models excel in extracting intricate patterns from vast datasets and adapting to diverse downstream tasks, ranging from natural language understanding and generation to specialized domains such as smart healthcare~\cite{cardenas2024autohealth,tang2024merit}, finance~\cite{wu2023bloomberggpt}, and intelligent transportation~\cite{tian2023vistagpt,hu2024agentscodriver}. However, due to immense parameter sizes, training LLMs from scratch is typically infeasible~\cite{hu2021lora}, rendering fine-tuning with private data for specific downstream tasks become mainstream. The workflow of fine-tuning LLMs typically follows a three-stage process: i) pre-training LLMs from scratch on extensive text corpora (e.g., Wikipedia) using substantial computing resources~(e.g., training  GPT3-1.3B model requires 64 Tesla V100 GPUs running for one week~\cite{yuan2022decentralized}); ii) fine-tuning pre-trained LLMs to adapt to specific downstream tasks; iii) deploying the fine-tuned LLMs on client devices to perform inference.

Unfortunately, privacy concerns often hinder the collaborative fine-tuning of LLMs across multiple parties, even when they share the same downstream task and could mutually benefit. For example, several freelance financial analysts may wish to collaboratively fine-tune an LLM for a special financial risk prediction task via their personal devices, but they are reluctant to share customer data~\cite{wu2023bloomberggpt}. Federated learning (FL)~\cite{mcmahan2017communication,hu2024accelerating} offers a promising solution to this privacy issue~\cite{shin2022fedbalancer, panchal2024flow} by enabling participants to fine-tune LLMs using local data without sharing the raw data~\cite{wang2024flora,zhang2024satfed}. However, fine-tuning LLMs via FL is non-trivial, particularly for resource-constrained client devices.  In typical configurations, client devices are equipped with low-cost commercial GPUs, such as the NVIDIA GeForce RTX 3050 in a laptop. While these GPUs are sufficient for conventional deep learning tasks such as object detection~\cite{dong2022wave} and image classification~\cite{lin2024efficient}, they can be overwhelmed by the immense computing workload of full parameter fine-tuning of LLMs within standard FL frameworks. To combat this issue, some recent studies~\cite{cai2023efficient,che2023federated} focus on integrating FL with parameter-efficient fine-tuning (PEFT) methods, such as Adapter~\cite{houlsby2019parameter} and low-rank adaptation (LoRA)~\cite{hu2021lora}. Despite these efforts, fine-tuning LLMs via FL on client devices remains prohibitively time- and resource-consuming. 
For instance, fine-tuning LLaMA 3-8B~\cite{touvron2023llama} requires over 20GB of GPU memory, which exceeds the capabilities of most client devices.

\begin{figure}[t] 
\centering
\includegraphics[width=.99\columnwidth]{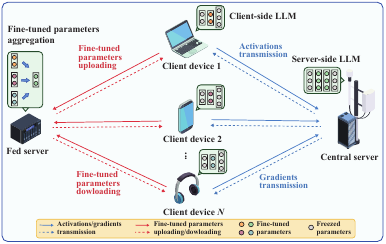}
\caption{A scenario of LLM SL via PEFT with client devices.}
\label{fig:llm_sl_teaser}
\end{figure}

Split learning (SL)~\cite{vepakomma2018split} has emerged as a compelling alternative capable of overcoming the weaknesses of FL. SL offloads the primary training workload to a central server via layer-wise model partitioning, alleviating the computing burden on resource-constrained client devices~\cite{lin2024adaptsfl}. Fig.~\ref{fig:llm_sl_teaser} illustrates how PEFT integrates with the state-of-the-art SL implementation, split federated learning (SFL)~\cite{thapa2022splitfed}, for LLM fine-tuning. The pre-trained LLM is split into two sub-models: client-side LLM deployed on the client devices and server-side LLM deployed on a central server. Each client device executes forward propagation on the client-side LLM and transmits activations to the central server. Then, the central server leverages the received activations to fine-tune the server-side LLMs and send gradients to client devices for client-side back-propagation. After the training is completed, the fed server aggregates the client-side fine-tuned parameters and redistributes them to client devices. This process repeats until the LLM reaches a pre-defined accuracy threshold. 

While integrating SL and PEFT techniques holds significant promise, its implementation encounters several critical challenges. First, the limited computing resources\footnote{In this paper, we use GPU memory footprint as a proxy metric to quantify computing resource availability~\cite{dhar2024empirical,xu2024edgellm,qu2025mobile}. This is because GPU memory footprint directly correlates with a model's computing demands, especially for LLM fine-tuning, where memory-intensive operations such as multi-head self-attention and large-scale matrix multiplications dominate. GPU memory footprint provides an effective proxy for evaluating the feasibility of model fine-tuning on resource-constrained client devices.} of client devices render fine-tuning all trainable parameters impractical, necessitating selectively fine-tuning a portion of trainable weights. However, this paradigm remains unexplored in the SL paradigm.
Second, the inherent heterogeneity of computing resources across client devices leads to a severe device unavailability problem\footnote{The device unavailability problem occurs when participating devices fail to perform local model training or data transmission due to disconnections, network instability, or insufficient computing resources, thus hindering global model updates and degrading training performance.}~\cite{gu2021fast1,theodoropoulos2024federated,yang2021characterizing}, ultimately degrading the training performance. Last but not least, the varying computing budgets across client devices result in structural discrepancies in fine-tuning adapters, making it infeasible to aggregate these adapters across diverse client devices. Empirical studies have been conducted to validate these challenges, as will be discussed in Section~\ref{sec:bg}.

To tackle these challenges, we propose a \underline{h}eterogeneous PEFT framework built on \underline{split} learning and \underline{LoRA} fine-tuning method, named \name. First,  in order to efficiently fine-tune the trainable weights under computing resource constraints, we design an important weight identification scheme to identify important weights based on their contributions to LLM training. Second, to accommodate the heterogeneous computing capabilities of edge
servers, we propose a dynamic rank and model splitting
configuration to dynamically adjust the decomposition ranks of LoRA adapters and the model split point to align with client devices' computing budgets. Finally, to seamlessly aggregate heterogeneous LoRA adapters, we develop a noise-free adapter aggregation that meticulously concatenates low-rank decomposition matrices to support heterogeneous adapter aggregation without introducing noise. 
The key contributions of this paper are summarized as follows:
\begin{itemize}
    \item To the best of our knowledge, \name is the first heterogeneous LLM fine-tuning framework that integrates SL and PEFT, enabling distributed collaborative learning across heterogeneous client devices.
    \item We devise an important weight identification scheme to quantify the contribution of trainable weights to training performance to prioritize important weights for fine-tuning.
    \item We propose an adaptive rank and model splitting configuration to adjust the decomposition ranks of LoRA adapters and the model split point to accommodate heterogeneous computing budgets of client devices.
    \item We develop a noise-free adapter aggregation that meticulously concatenates low-rank decomposition matrices to support heterogeneous adapter aggregation without introducing additional noise. 
    \item We empirically evaluate \name with extensive experiments. The results demonstrate that \name outperforms state-of-the-art benchmarks, significantly in training accuracy and convergence speed. 
\end{itemize}

The rest of this paper is organized as follows. Section~\ref{sec:bg} introduces the background and motivation. Section~\ref{sec:design}
elaborates on the system design of \name. Section~\ref{sec:impl} presents the system implementation, followed by the performance evaluation in Section~\ref{sec:eval}. Related works and technical limitations are discussed in Section~\ref{sec:rw}. Finally, conclusions and future remarks are presented in Section~\ref{sec:con}.

\begin{figure}[t]
\centering
\includegraphics[width=0.8\linewidth]{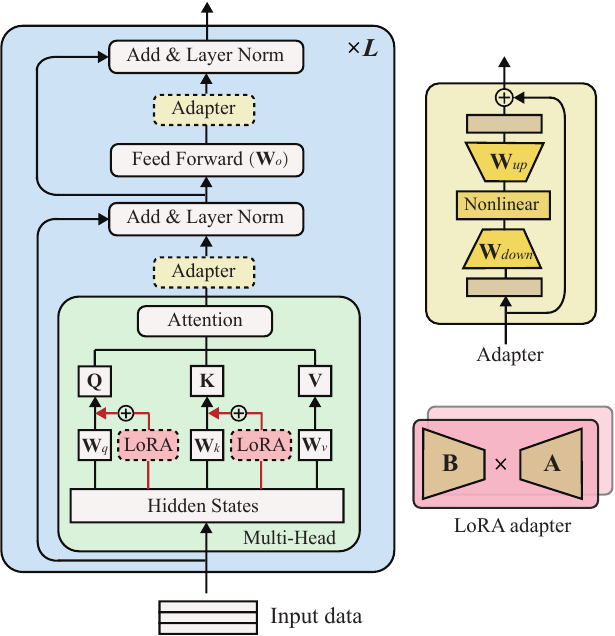}
\caption{An illustration of Adapter~\cite{houlsby2019parameter} and LoRA~\cite{hu2021lora} fine-tuning methods, where ${{\bf{W}}_q}$, ${{\bf{W}}_k}$, ${{\bf{W}}_v}$, and ${{\bf{W}}_o}$ are trainable weights~\cite{lin2023pushing}.}
\label{fig:eg_fine_tune}
\end{figure}

\section{Background and Motivation} \label{sec:bg}
In this section, we begin by introducing the background of PEFT for LLM SL. Then, we conduct extensive pilot measurement studies to investigate the design challenges of SL on LLM PEFT, serving as the foundation for the design motivation behind \name.

\subsection{Injecting PEFT into Split Learning for LLMs} \label{ssec:bg:peft}

The prevalent LLMs typically consist of billions or even hundreds of billions of parameters. The immense scale of LLMs demands substantial computing resources, rendering full-parameter fine-tuning (FT)~\cite{kenton2019bert} (i.e., initializing the model with pre-trained weights and updating all parameters to adapt to various downstream tasks) of LLMs infeasible, especially on resource-constrained client devices. While the SL can offload primary computing burdens to central servers via model partitioning~\cite{lin2023pushing}, fine-tuning client-side LLMs still leads to intolerable latency. For instance, the client-side LLM composed of the first 10 Transformer blocks from the LLama-2-7B model~\cite{touvron2023llama} contains 8.75\!~billion parameters—60 times more than the parameter count of traditional models like VGG-16~\cite{simonyan2014very}, which has only 139 million parameters.


To address these challenges, PEFT offers a promising solution by fine-tuning only a small portion of LLM parameters, significantly reducing the computing cost. The current two widely adopted PEFT methods are Adapter~\cite{houlsby2019parameter} and LoRA~\cite{hu2021lora}. As shown in Fig.~\ref{fig:eg_fine_tune}, Adapter works by inserting a few additional trainable layers into the model backbone~\cite{houlsby2019parameter} and only fine-tuning
those layers in a pre-trained model. In contrast, LoRA concatenates two trainable low-rank decomposition matrices in parallel to weights~\cite{hu2021lora} but freezes all weights of a pre-trained model. For LoRA fine-tuning, the model update can be expressed as $\mathbf{W}=\mathbf{W}_0 + \Delta \mathbf{W}$. Here, $\Delta \mathbf{W} = \mathbf{B}\mathbf{A}$  denote incremental model update, $\mathbf{B} \in \mathbb{R}^{d_i \times r}$ and  $\mathbf{A}\in \mathbb{R}^{r \times d_o}$ are low-rank decomposition matrices. The rank is $r\ll \min{(d_i, d_o)}$ (e.g., $r = 4$ when $d_i = d_o = 1024$), and dimensions of input and output of $\bf B$ and $\bf A$ are $d_i$ and $d_o$, respectively. The LoRA adapter is the cascade structure composed of $\mathbf{B}$ and $\mathbf{A}$. Compared to Adapter, LoRA offers two distinct advantages: First, LoRA requires lower computing costs by updating only two low-rank matrices with few parameters. Second, LoRA can support LLM fine-tuning without changing their backbone architectures, showcasing better scalability and flexibility. Several studies~\cite{hu2021lora, hu2023llm, sheng2023s} have demonstrated the superior performance of LoRA over Adapter, and hence we focus exclusively on LoRA in the following discussions and designs.

\begin{figure}[t]
\centering
\subfloat[{Computing overhead}\label{fig:llm_train:computing}]{
\includegraphics[width=0.48\columnwidth]{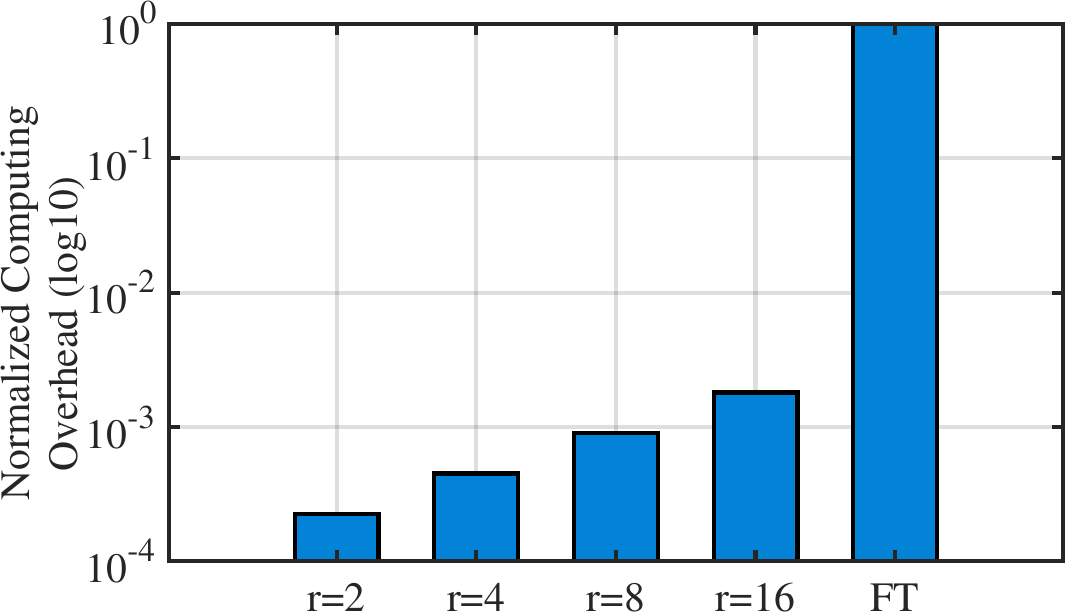}
}
\subfloat[{Communication overhead}\label{fig:llm_train:communication}]
{
\includegraphics[width=0.48\columnwidth]{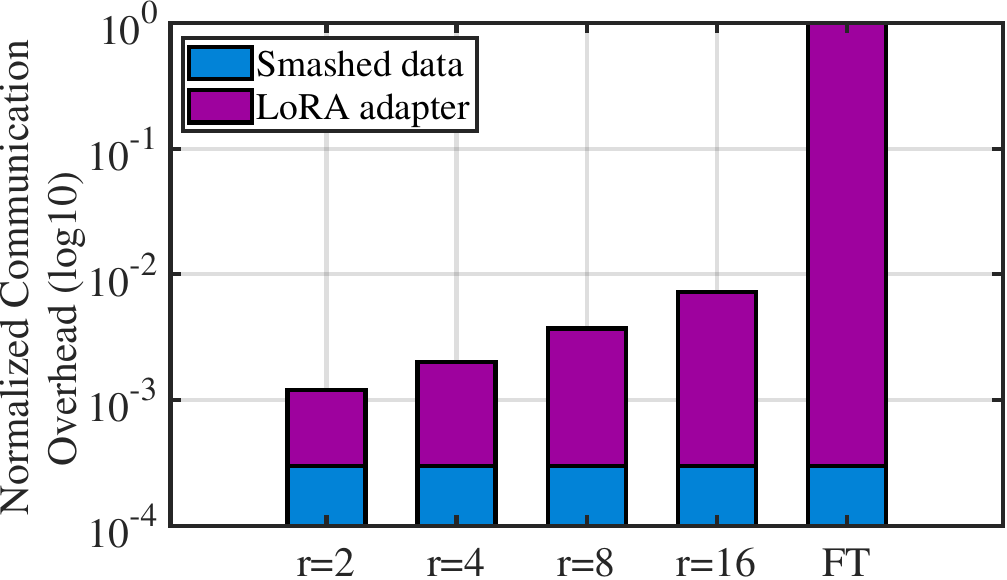}
}
\caption{The normalized computing and communication overheads of LoRA and FT on LLaMA-2-7B, where $r$ is the decomposition rank of LoRA adapter.}
\label{fig:llm_training}
 \vspace{-2ex}
\end{figure}

To better understand the computing and communication bottlenecks in LLM SL, we integrate the LoRA fine-tuning technique into one of the most representative SL frameworks, SFL~\cite{thapa2022splitfed}, and take the well-known LLaMA-2-7B model as the global model. We select the E2E dataset~\cite{novikova2017e2e} as the training dataset. Unless otherwise specified, this experimental setup remains the same throughout the whole Section II. Fig.~\ref{fig:llm_train:computing} and \ref{fig:llm_train:communication} present the normalized 
computing and communication overhead\footnote{The communication overhead consists of two components: the smashed data (i.e., activations and gradients) exchange overhead between client devices and the central server for model fine-tuning, and the transmission overhead of LoRA adapters for model aggregation.} for LoRA and FT. The results reveal that the computing and communication overhead of FT is more than 500 and 100 times that
of LoRA. Apparently, instead of the FT, PEFT can significantly mitigate the computing and communication overhead of LLM SL.
However, we still meet three key challenges in integrating PEFT into LLM SL, which will be discussed in the following sections.

\subsection{Limited Computing Resources} \label{ssec:bg:weight}

While LoRA significantly reduces the computing overhead for fine-tuning LLMs~(see Section~\ref{ssec:bg:peft}), configuring LoRA adapters to all trainable weights on resource-constrained client devices is still impractical as LLM scale up~\cite{sun2024federated,hu2021lora,dettmers2024qlora}. In SL, this limitation also holds for the central server, which needs to handle heavy workloads offloaded from multiple client devices. Despite its more powerful computing capability, the central server still struggles to accommodate LoRA configurations for all trainable weights as the number of served client devices increases~\cite{pal2021server,lin2024efficient}. Therefore, it is imperative to selectively fine-tune trainable weights under resource constraints.

\begin{figure}[t!]
  \centering
  \subfloat[{Index}\label{fig:diff_config:layers}]{
    \includegraphics[width=0.455\columnwidth]{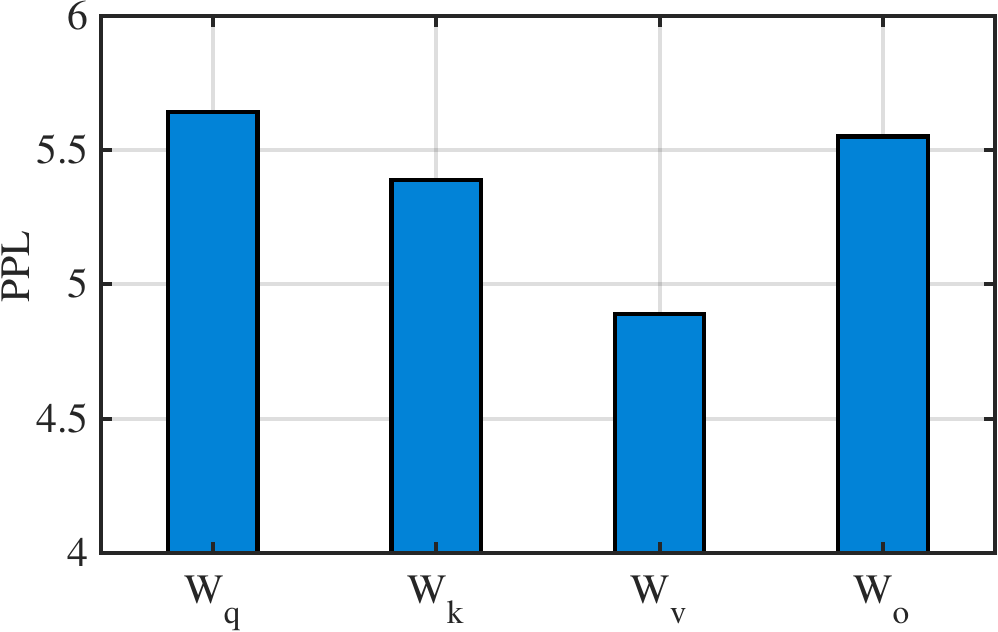}
  }
  \subfloat[{Quantity} \label{fig:diff_config:chan}]{
  \includegraphics[width=0.451\columnwidth]{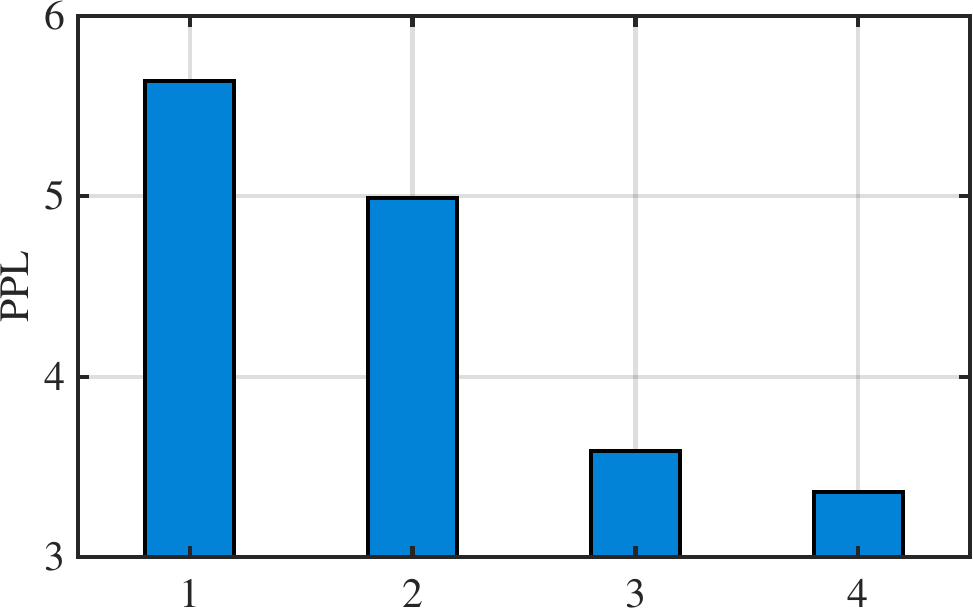}
  }
  \caption{The training performance with different trainable weights on LLaMA-2-7B, where the trainable weights combinations corresponding to quantities 1, 2, 3, and 4 are $\left\{ {{{\bf{W}}_q}} \right\}$, $\left\{ {{{\bf{W}}_q},{{\bf{W}}_k}} \right\}$, $\left\{ {{{\bf{W}}_q},{{\bf{W}}_k},{{\bf{W}}_v}} \right\}$, and $\left\{ {{{\bf{W}}_q},{{\bf{W}}_k},{{\bf{W}}_v},{{\bf{W}}_o}} \right\}$, respectively.}
  \label{fig:diff_config}
   \vspace{-2ex}
\end{figure}

\begin{figure}[t]
  \centering
  \subfloat[{PPL}\label{fig:important_ppl_20round}]
  {
    \includegraphics[width=0.48\columnwidth]{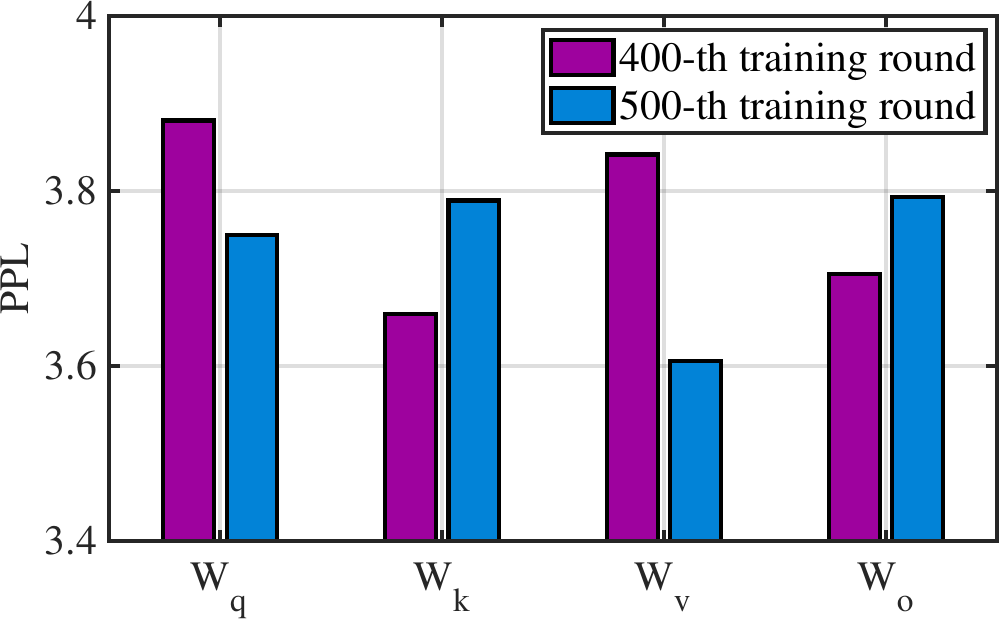}
  }
  \subfloat[{STD of PPL}\label{fig:important_std}]
  {
    \includegraphics[width=0.48\columnwidth ]{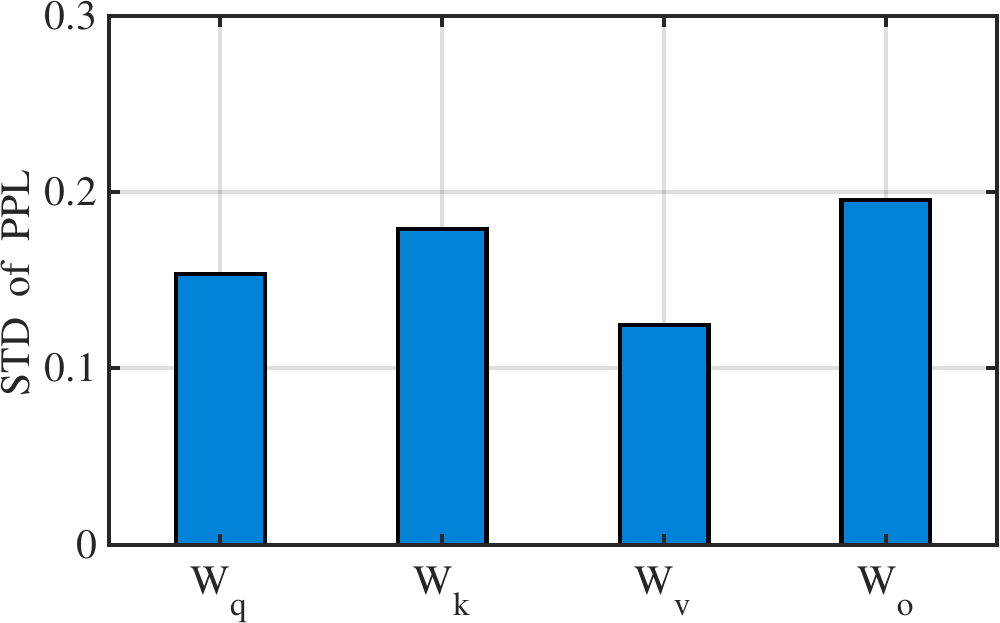}
  }
  \caption{The PPL at the $400$-th and $500$-th training rounds and the STDs of the PPL across 100 rounds on LLaMA-2-7B. }
  \label{fig:weight_dynamic}
   \vspace{-2.3ex}
\end{figure}

To understand how the selection of trainable weights impacts the training performance of SL, we conduct two motivating experiments. Perplexity~(PPL) serves as the performance metric to evaluate the training performance of the model, with a smaller PPL indicating better performance.  First, we fine-tune the LLama-2-7B model for individual trainable weights (i.e., ${{\bf{W}}_q}$, ${{\bf{W}}_k}$, ${{\bf{W}}_v}$, and ${{\bf{W}}_o}$ in Fig.~\ref{fig:eg_fine_tune}) and combinations with varying numbers of trainable weights. Fig.~\ref{fig:diff_config:layers} illustrates that different trainable weights contribute unequally to the model's training performance, with some weights yielding more significant improvements than others. Fig.~\ref{fig:diff_config:chan} demonstrates that more trainable weights achieve better training performance, but this comes at the cost of higher computing overhead. 
Second, we configure LoRA adapters for four trainable weights across two distinct training rounds to investigate the dynamic impact of trainable weights on model performance. Fig.~\ref{fig:important_ppl_20round} shows the PPL of each trainable weight, and Fig.~\ref{fig:important_std} reports the standard deviation (STD) of PPL. The results indicate that the contribution of trainable weights to training performance varies throughout training, rendering the fixed fine-tuning scheme infeasible. Therefore, it is paramount to dynamically select the combination of trainable weights for LoRA adapter configuration.

\subsection{Severe Device Unavailability Problem} \label{ssec:bg:straggler}

Current state-of-the-art SL frameworks~\cite{vepakomma2018split,thapa2022splitfed} typically assume sufficient and homogeneous computing resources across client devices for synchronous local model training, thereby leading to good training performance. However, in practice, the computing capabilities of client devices often vary due to differences in hardware architecture~\cite{jhang2021challenges} and the priority of on-demand running programs~\cite{zhang2023resource}. Client devices with weak computing power may fail to complete local model training, resulting in a severe device unavailability problem~\cite{zhang2023timelyfl, horvath2021fjord, ouyang2022clusterfl} and thus severely degrading training efficiency and performance. To gain a deeper understanding of the impact of computing heterogeneity on LLM SL, we conduct two motivating experiments. Consistent with Section~\ref{ssec:bg:weight}, we integrate the LoRA fine-tuning technique into SFL~\cite{thapa2022splitfed} and evaluate its performance on LLaMA-2-7B and GPT-2-L. 

Fig.~\ref{fig:straggler:comp} and Fig.~\ref{fig:straggler:comm} show the converged accuracy and time on LLaMA-2-7B and GPT-2-L under varying device unavailability rates\footnote{The device unavailability rate is defined as the proportion of devices incapable of local training due to computing limitations, which is calculated as the number of devices unable to complete local training divided by the total number of participating devices.}. The results demonstrate that the converged accuracy degrades by 0.23 and 0.21 PPL, and convergence speed slows down by 42\% and 40\% on LLaMA-2-7B and GPT-2-L, respectively, as the device unavailability rate increases from 0\% to 50\%. Therefore, it is essential to develop an efficient SL fine-tuning framework capable of accommodating heterogeneous computing resources of client devices. 

\begin{figure}[t]
  \centering
  \subfloat[{Decomposition rank}\label{fig:rank_gpt2}]
  {
    \includegraphics[width=0.503\columnwidth]{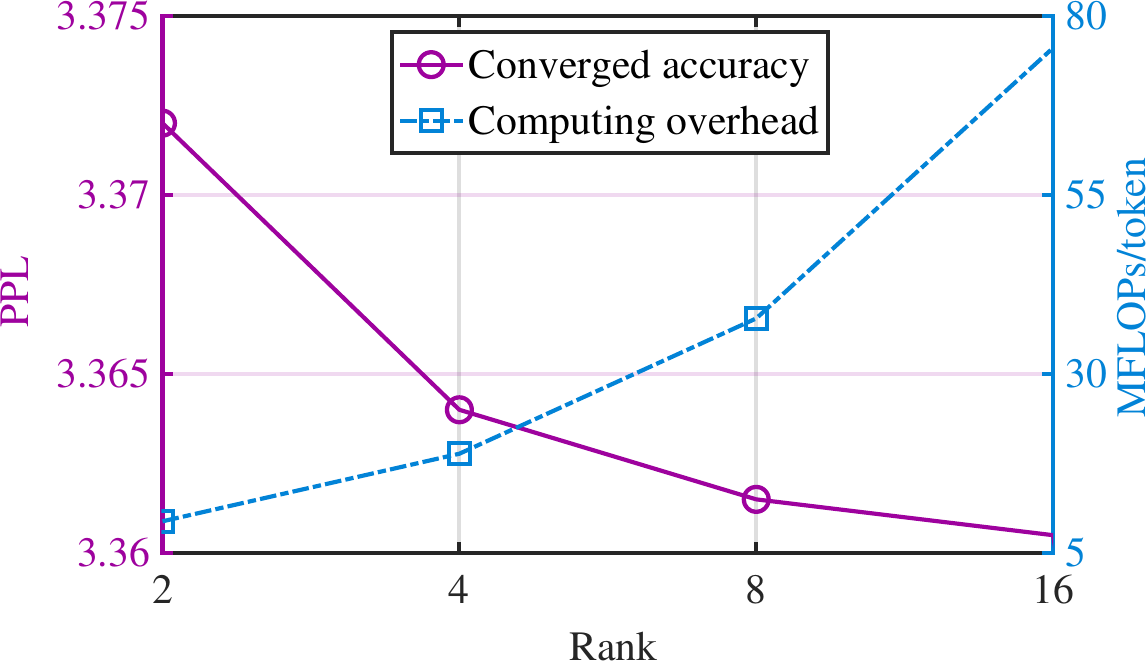}
  }
  \subfloat[{Model split point}\label{fig:cut_gpt2}]
  {
    \includegraphics[width=0.480\columnwidth]{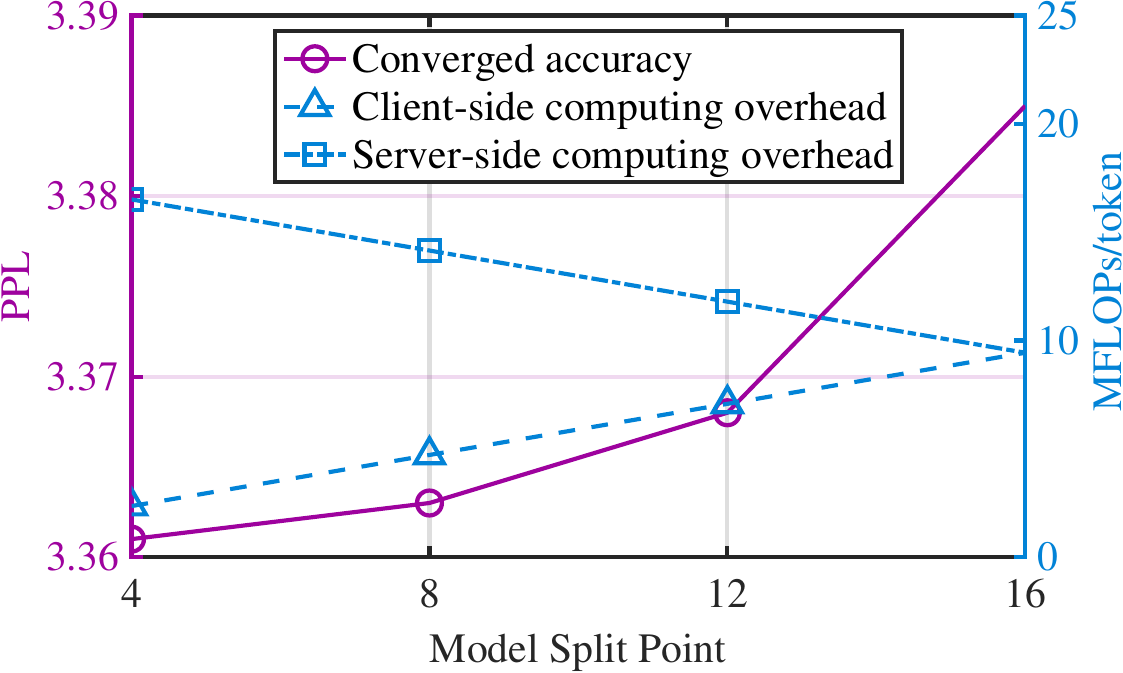}
  }
  \caption{The converged accuracy and computing overhead versus decomposition rank and model split point on LLaMA-2-7B.}
  \label{fig:batch_and_rank}
   \vspace{-2ex}
\end{figure}

To combat the computing heterogeneity issue, customizing key parameters in LLM SL is critical. For LLM SL, the decomposition ranks of LoRA adapters and the model split point are pivotal for training performance and computing costs. The rank dictates the number of trainable parameters in each weight for fune-tuning~\cite{sheng2023s, li2024caraserve}, while the model split point determines the distribution of computing load between client devices and a central server~\cite{vepakomma2018split,thapa2022splitfed,lin2023pushing}. To investigate the impact of these two parameters on LLM SL, we fine-tune LLaMA-2-7B with varying decomposition rank and model split point configurations, as shown in Fig.~\ref{fig:batch_and_rank}. The model split point $j$ is defined as the model connection point between the $j$-th and $(j+1)$-th transformer block. It is seen from Fig.~\ref{fig:rank_gpt2} that increasing the decomposition rank improves training but incurs higher computing costs. Fig.~\ref{fig:cut_gpt2} illustrates that model split point trade-offs the computing load between the client devices and central server while affecting training performance. Therefore, it is vital to customize the optimal decomposition rank and model split point for each client device under heterogeneous computing constraints.

\begin{figure}[t!]
  \centering
    \subfloat[{Converged accuracy}\label{fig:straggler:comp}]{
    \includegraphics[width=0.466\columnwidth]{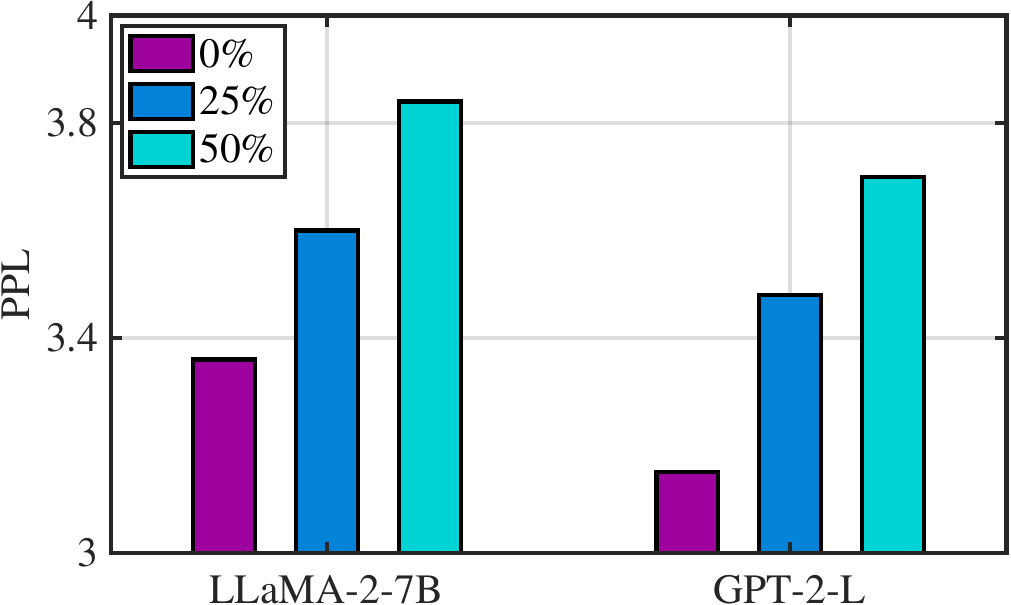}
  }
    \subfloat[{Converged time}\label{fig:straggler:comm}]
  { \includegraphics[width=0.475\columnwidth]{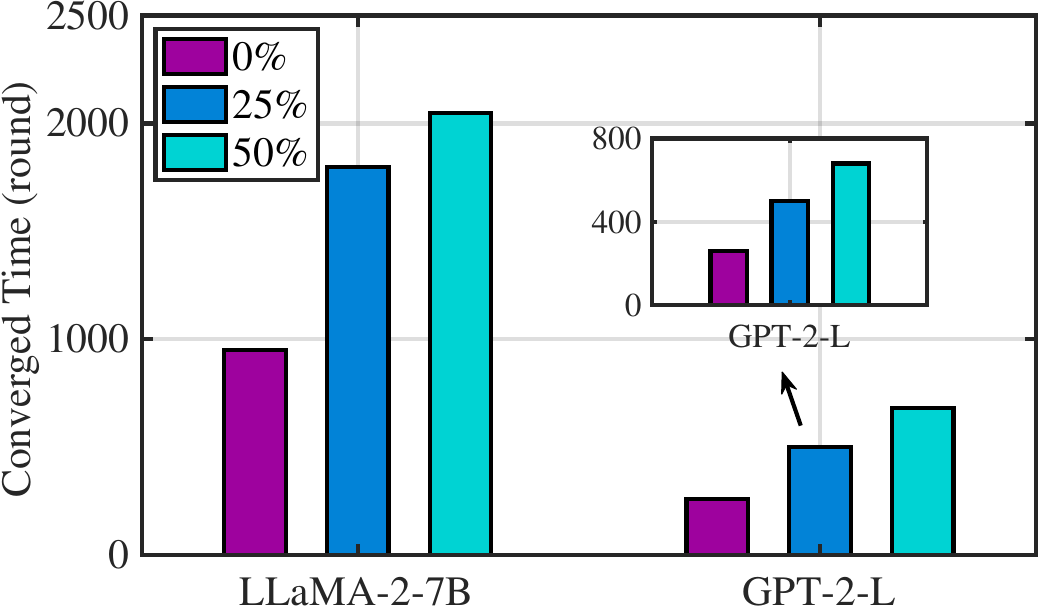}
  }
  \caption{The converged accuracy and time on LLaMA-2-7B and GPT-2-L under varying device unavailability rates.}
  \label{fig:straggler}
\end{figure}

\begin{figure}[t]
\centering
\includegraphics[width=0.76\linewidth]{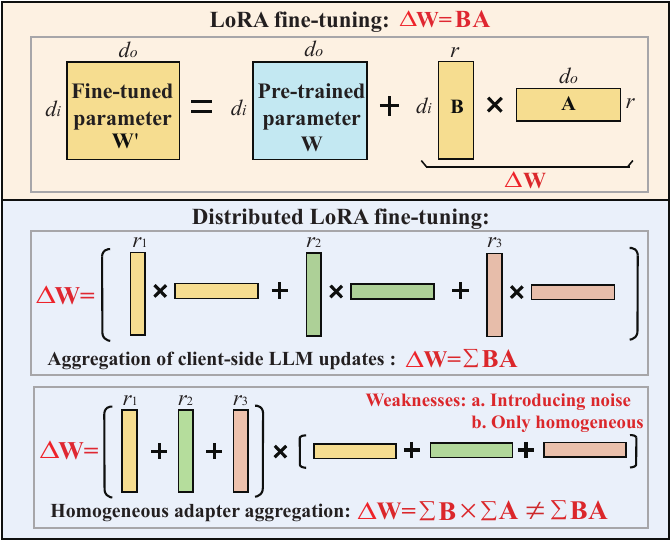}
\caption{The comparison of the state-of-the-art homogeneous aggregation scheme~\cite{zhang2024towards} with the aggregation of client-side LLM updates, where $\mathbf{B}$ and $\mathbf{A}$ are low-rank decomposition matrices.}
\label{fig:heter_lora_aggre_motivation}
\end{figure}

\subsection{Heterogenous Adapter Aggregation} \label{ssec:bg:heter_adapter_agg}

Recall Fig~\ref{fig:llm_sl_teaser}, the fed server is responsible for periodically aggregating LoRA adapters from all client devices to consolidate client-side LLM updates in LLM SL. As illustrated in Fig.~\ref{fig:heter_lora_aggre_motivation}, the current state-of-the-art aggregation scheme~\cite{zhang2024towards} assumes homogeneity in LoRA adapters across client devices. It first averages the low-rank decomposition matrices across client devices and then multiplies them to represent the aggregated client-side LLM update. However, this scheme is mathematically inconsistent with the aggregation of client-side LLM updates, which averages the product of the decomposition matrices across all client devices. This inconsistency inevitably causes mathematical errors and introduces additional noise into the adapter aggregation process. As discussed in Section~\ref{ssec:bg:straggler}, the inherent computing heterogeneity across client devices results in varying decomposition ranks for LoRA adapters across client devices. This dimensional mismatch between LoRA adapters renders the homogeneous adapter aggregation unsuitable, as it fails to accommodate the structural discrepancies across LoRA adapters. A more sophisticated adapter aggregation scheme is needed—one that can accommodate the heterogeneous nature of LoRA adapters while avoiding additional noise in the adapter aggregation process.

\section{System Design} \label{sec:design}

\subsection{Overview}
\begin{figure}[t]
\centering
\includegraphics[width=\linewidth]{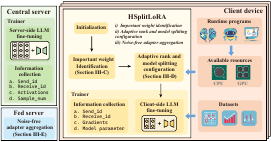}
\caption{An overview of \name architecture.}
\label{fig:overview}
\end{figure}

To combat address the above-mentioned challenges, we propose \name, a heterogeneous PEFT framework built on SFL~\cite{thapa2022splitfed} and LoRA fine-tuning method~\cite{hu2021lora}, as illustrated in Fig.~\ref{fig:overview}. \name consists of the following three meticulously designed components:

\begin{itemize}
    \item To facilitate efficient fine-tuning under computing constraints, we design an \textit{important weight identification} (Section~\ref{ssec:import_identify}) to identify the contribution of each trainable weight to training performance, so as to prioritize important weights for fine-tuning.
    \item To overcome the device unavailability issue, we propose an \textit{adaptive rank and model splitting configuration} (Section~\ref{ssec:dynamic_split_rank}), which adjusts the decomposition ranks of LoRA adapters and model split point based on heterogeneous computing budgets of client devices to improve training efficiency.
    \item To efficiently aggregate heterogeneous adapters, we develop a \textit{noise-free adapter aggregation} (Section~\ref{ssec:aggregate}) that meticulously concatenates low-rank decomposition matrices to eliminate structural discrepancies across adapters, achieving noise-free heterogeneous adapter aggregation.
\end{itemize}

The training workflow of \name for each training round follows three steps: i) Each client device utilizes the important weight identification to determine the fine-tuning priority of trainable weights; ii) client devices conduct adaptive rank and model splitting configuration based on computing budgets, and then collaborate with the central server for LLM fine-tuning via activations/gradients exchange; and iii) fed server employs noise-free adapter aggregation to aggregate heterogeneous client-side adapters and distribute aggregated LoRA adapters to all client devices for the next training round. It is noted that the last step can be conducted less often. \name is built upon a common SL framework~\cite{thapa2022splitfed}, where aggregation refers to a weighted average of several adapters (similar to FedAvg~\cite{mcmahan2017communication}) but only for adapters trained on the client side.

\subsection{System Model} \label{ssec:lora}

\begin{figure}
\centering
\includegraphics[width=1\linewidth]{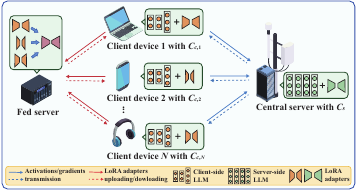}
\caption{Integrating LoRA adapters into LLM SL.}
\label{Scenario}
 \vspace{-1ex}
\end{figure}

In this section, we model the split fine-tuning of LLM on client devices with heterogeneous computing budgets, providing a theoretical foundation for the design of \name. As shown in Fig.~\ref{Scenario}, we consider a typical scenario of \name over an edge computing system, consisting of three components:

\begin{itemize}
\item \textbf{Client device:} 
The set of participating client devices is denoted by $\mathcal{N} = \left\{ {1,2,...,N} \right\}$, where $N$ is the total number of client devices. The computing budget of the $n$-th client device is $C_{c, n}$, while the client-side LLM on the $n$-th client device is represented as ${\bf W}_{c, n}$. The local dataset residing on the $n$-th client device with $|{\mathcal{D}_n}|$ data samples is denoted by ${\mathcal{D}_n} = \left\{ {{{\bf{x}}_{n,k}},{y_{n,k}}} \right\}_{k = 1}^{{|{\mathcal{D}_n}|}}$, where ${{\bf{x}}_{n,k}}$  and ${{{y}}_{n,k}} $ are the $k$-th input data and its corresponding label in ${\mathcal{D}_n}$, respectively.
Consequently, the total dataset across all client devices is $ \mathcal{D} = {\textstyle \bigcup_{n=1}^{N} \mathcal{D}_{n}}$.
\item \textbf{Central server:} The central server is a powerful computing entity tasked with fine-tuning server-side LLM.  The computing budget of the central server is denoted by $C_{s}$ and the server-side LLM is ${\bf W}_{s}$.
\item \textbf{Fed server:} The fed server is an entity responsible for adapter aggregation, periodically aggregating client-side LoRA adapters from all participating client devices. 
\end{itemize}

The global model of the $n$-th client device is represented as ${{\bf{W}}_n} = \left[ {{{\bf{W}}_s};{{\bf{W}}_{c, n}}} \right]$. The local loss function of the $n$-th client device is ${L}_{n} (\mathbf{W}_n) = \frac{1}{\left | \mathcal{D}_{n} \right | }  \sum_{k=1}^{\left | \mathcal{D}_{n} \right |} L_{n,k} (\mathbf{x}_{n,k}, y_{n,k}; \mathbf{W}_n)$, where $L_{n,k} (\mathbf{x}_{n,k}, y_{n,k}; \mathbf{W}_n)$ denotes the sample-wise loss function of the $k$-th data sample in $\mathcal{D}_{n}$. The objective of SL is to find the optimal global model that achieves good performance across all participating client devices, which can be formulated to minimize the finite-sum nonconvex global loss function:
\begin{equation} \label{eq:fl_formu}
    \min_{\mathbf{W}_g} {L(\mathbf{W}_g)} = \min_{\mathbf{W}_g}\sum_{n=1}^{N}\frac{\left|\mathcal{D}_{n} \right|}{\left| \mathcal{D} \right | }{L_{n}(\mathbf{W}_g)},
\end{equation}
where ${{\bf{W}}_g} = \sum\limits_{n = 1}^N {\frac{{\left| {{{\mathcal D}_n}} \right|}}{{\left| {\mathcal D} \right|}}{{\bf{W}}_n}} $.

As discussed in Section~\ref{ssec:bg:straggler}, decomposition rank and model split point can significantly impact the LoRA fine-tuning performance. Therefore, we reformulate Eqn.~\eqref{eq:fl_formu} as
\begin{subequations}\label{optimization problem}
\begin{alignat}{2}
\min_{S, \mathcal{R}_{n}, \mathcal{W}_{n}} \quad & \sum_{n=1}^{N}\frac{\left|\mathcal{D}_{n} \right|}{\left| \mathcal{D} \right | }{L_{n}(\mathbf{W}_g | S, \mathcal{R}_{c, n}, \mathcal{R}_{s}, \mathcal{W}_{c, n}, \mathcal{W}_{s})}, & \tag{2} \label{eq:opt:2a} \\
\mbox{s.t.}\quad
& S \in \mathcal{S}, &\tag{2a} \label{eq:opt:2b} \\
& \mathcal{W}_{c, n}, \mathcal{W}_{s} \subseteq \{\mathbf{W}_q,\mathbf{W}_k,\mathbf{W}_v,\mathbf{W}_o \}, &\tag{2b} \label{eq:opt:2c} \\
& {r}_{c, n}^m, {r}_{s}^m  \in  \mathcal{Q}, &\tag{2c} \label{eq:opt:2d} \\
& { f_{c}(\mathcal{W}_{c, n}, \mathcal{R}_{c, n})\le C_{c, n}}, &\tag{2d}\label{eq:opt:2e}\\
& { f_{s}(\mathcal{W}_{s}, \mathcal{R}_{s})\le C_{s}}, &\tag{2e}\label{eq:opt:2f}
\end{alignat}
\end{subequations}
where $S$ is the model split point selection decision, $\mathcal{W}_{c,n}$ and $\mathcal{W}_s$ represent selected combinations of trainable weights for the $n$-th client device and central server, $\mathcal{R}_{c,n} = \{r_{c,n}^1, r_{c,n}^2, \ldots, r_{c,n}^M\}$ and $\mathcal{R}_s = \{r_s^1, r_s^2, \ldots, r_s^M\}$ denote sets of decomposition ranks of $\mathcal{W}_{c,n}$ and $\mathcal{W}_s$. Here, $r_{c,n}^m$ and $r_s^m$ represent the decomposition rank of the $m$-th trainable weight for the $n$-th client device and central server\footnote{If the $m$-th trainable weight of the $n$-th client device/central server is not selected for LoRA fine-tuning, its decomposition rank $r_{c,n}^m/r_s^m$ is set to zero.}, and $M$ is the total number of trainable weights\footnote{Recalling Fig. 2, for LoRA fine-tuning, trainable weights are query, key, value, and feed-forward layers, resulting in $M=4$.}.

Eqn.~(2a) ensures that the model split point is selected from a predefined set $S$. Eqn.~(2b) specifies that the trainable weights are chosen from the query, key, value, and feed-forward layers of Transformer architecture (see Fig.~2). Eqn.~(2c) guarantees that the decomposition rank of the LoRA adapter is selected from a predefined rank set $\mathcal{Q}$. Eqn.~(2d) and Eqn.~(2e) represent the computing budgets of the $n$-th client device and central server, where $f_c(\cdot)$ and $f_s(\cdot)$ map the structure of trainable weights with LoRA adapters into the corresponding computing budget. The optimization problem in Eqn.~(2) is a mixed-integer linear programming (MILP) problem, which is typically NP-hard. To solve this problem, we propose {HSplitLoRA}, a heterogeneous PEFT framework elaborated in Sections~III-C to III-E.

\subsection{Important Weight Identification} \label{ssec:import_identify}

As explained in Section~\ref{ssec:bg:weight}, limited computing budgets of client devices and central server necessitate selective fine-tuning of trainable weights. To this end, we design an important weight identification scheme to identify and predict important weights (i.e., trainable weights with significant contributions to training performance) for the next training round, enabling efficient fine-tuning under constrained computing budgets. The scheme comprises two key components: a weight importance metric and dynamic weight identification.


{\bf a) Weight Importance Metric.} Existing method for determining weight importance typically rely on single-dimensional metrics~\cite{lecun2015deep,hanin2018neural,blalock2020state,tan2019efficientnet}, such as gradients~\cite{lecun2015deep,hanin2018neural}, weight magnitudes~\cite{blalock2020state}, or computational complexity~\cite{tan2019efficientnet}. These methods fail to capture the multifaceted nature of weight importance. For instance, gradients may inadequately reflect the relative significance of weights in the model~\cite{lecun2015deep,hanin2018neural}, while weight magnitudes may ignore the sensitivity to training dynamics~\cite{blalock2020state}. Furthermore, computing costs of different weights vary significantly, and overlooking this may result in overestimating the importance of computation-intensive weights, hence undermining the effectiveness of fine-tuning. This limitation underscores the need for a holistic metric that accounts for both optimization impact and computing demands.

We propose a new metric, named \underline{r}esource-\underline{n}ormalized \underline{g}radient-\underline{w}eight \underline{p}roduct (RNGWP), to evaluate weight importance by capturing the interplay between weights and gradients, as well as associated computing costs. For an arbitrary trainable weight $\bf{W}$, RNGWP is given by
\begin{equation}\label{eqn:average_gra_wei}
\Theta \left( {\bf{W}} \right) = \frac{{\sum\limits_{j = 1}^J {\left| {{w_j}{\nabla _{{w_j}}}L\left( {\bf{W}} \right)} \right|} }}{{C\left( {\bf{W}} \right)}},
\end{equation}
where ${w_j}$ and ${{\nabla _{{w_j}}}L\left( {\bf{W}} \right)}$ denote the $j$-th parameter of the model weight ${\bf{W}}$ and its corresponding gradient, respectively; $J$ is the total number of model parameters in $\bf W$; and ${C\left( {\bf{W}} \right)}$ denotes the fine-tuning cost\footnote{We utilize GPU memory footprint to quantify the fine-tuning cost of the trainable weight~\cite{dhar2024empirical,xu2024edgellm,qu2025mobile}, as it represents a critical resource constraint in fine-tuning LLM on client devices. Specifically, the GPU memory footprint comprises model parameters, activations and activations' gradients, optimizer state (depending on the choice of the optimizer, e.g. SGD, Momentum, and Adam), which can be systematically estimated following the methodology in~\cite{lin2024hierarchical,yeung2021horus}.} of trainable weight $\bf W$.

{\bf Remark:} The numerator of RNGWP, $\left| {{w_j}{\nabla _{{w_j}}}L\left( {\bf{W}} \right)} \right|$, represents the magnitude of the gradient-weight product, capturing both static and dynamic contributions of weights to training performance: The weight magnitude reflects its inherent significance in the model (i.e., static contribution), while the gradient quantifies the weight's sensitivity to the training loss (i.e., dynamic contribution). This design prevents the overestimation of large weights with negligible gradients and the over-prioritization of small weights with steep gradients. To accommodate varying computing budgets across weights, the gradient-weight product is normalized by the denominator ${C\left( {\bf{W}} \right)}$. This normalization ensures that fine-tuning prioritizes weights with high importance relative to their computing costs. 

\begin{figure}[t]
  \centering
  \subfloat[PPL\label{fig:moti_cur_his_accuracy}]{
    \includegraphics[width=0.475\columnwidth]{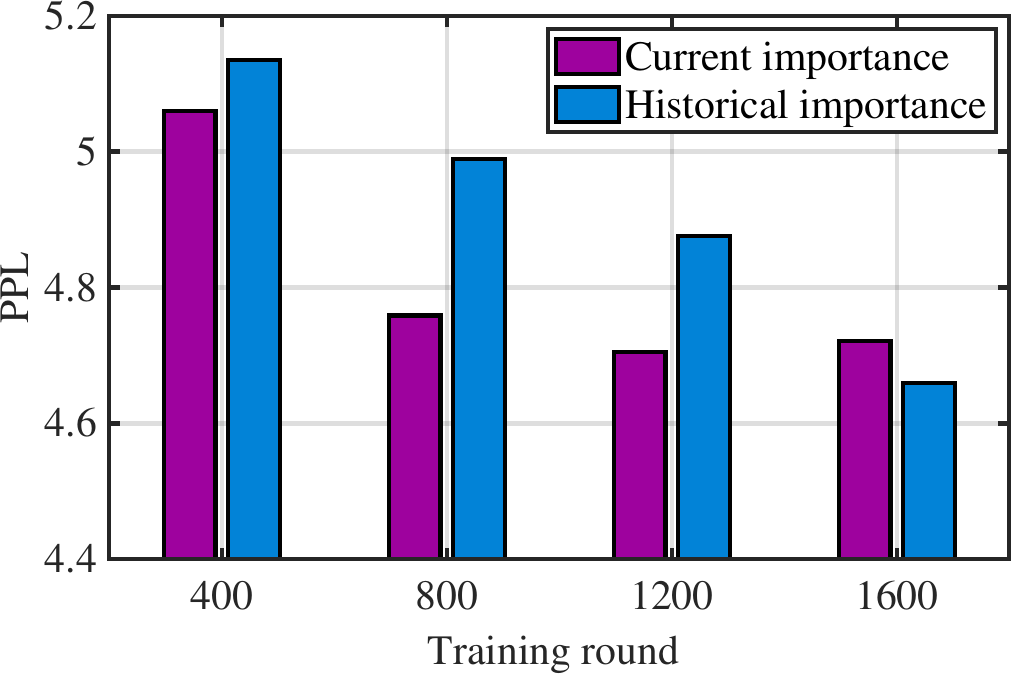}
  }
  \subfloat[STD of PPL\label{fig:moti_cur_his_std}]
  {
    \includegraphics[width=0.475\columnwidth]{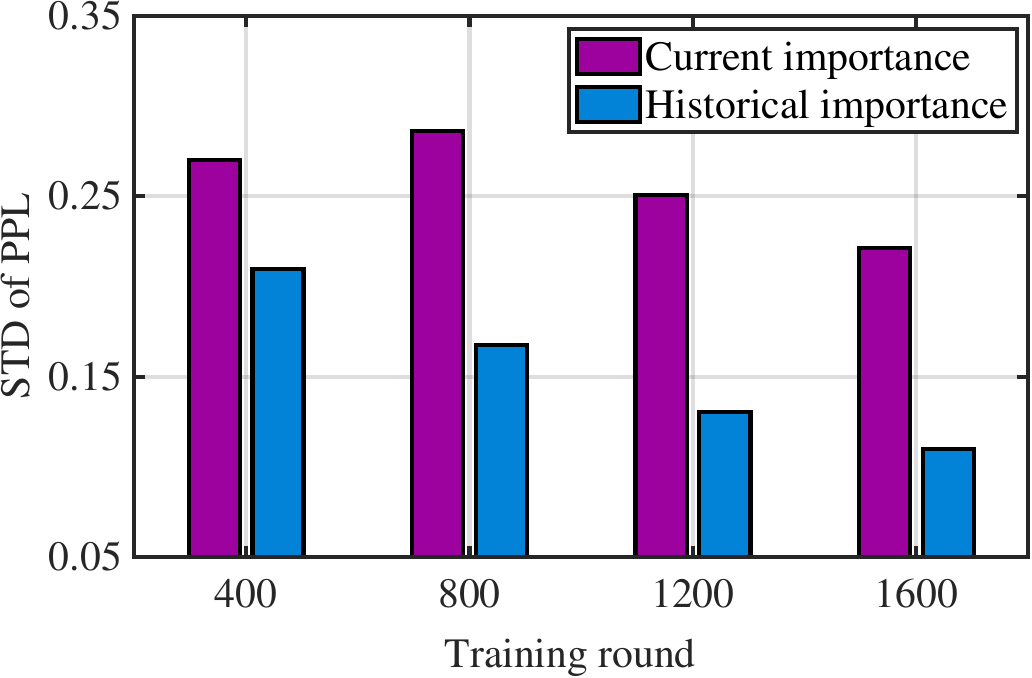}
  }
  \caption{The PPL and its STDs of fine-tuning a single weight with the highest current and historical importance on LLaMA-2-7B.}
  \label{fig:important_metric_compare}
 \vspace{-2ex}
\end{figure}

{\bf b) Dynamic Weight Identification.} As LLM fine-tuning progresses, the weight importance fluctuates due to changes in model parameters and computing budgets across client devices. 
While the current importance (i.e., the weight importance at the current training round) captures immediate importance fluctuations, relying solely on it is insufficient. LLM fine-tuning is prone to short-term noise or local optima{~\cite{zhang2023glm,wang2024flora}}, and focusing exclusively on current importance risks overfitting, compromising model convergence and fine-tuning performance{~\cite{wu2024fedbiot,wang2024twostage}}. In contrast, historical importance (i.e., the cumulative weight importance over several previous training rounds) reflects long-term trends of weight importance across training rounds{~\cite{kotha2024understanding,xu2024dofit}}, enabling the model to resist noise and remain stable model convergence.

To better understand the impact of current and historical importance on training performance, we fine-tune a single weight with the highest current and historical importance on LLaMA-2-7B. Fig.~\ref{fig:important_metric_compare} shows that fine-tuning with the current importance achieves rapid PPL reduction in the early stages of training, but exhibits significantly higher standard deviation (STD) than the counterpart with historical importance. In contrast, fine-tuning with the historical importance demonstrates more stable convergence, ultimately achieving superior training performance with lower STD. These results reveal a trade-off: current importance enables rapid adaptation but lacks convergence stability, whereas historical importance improves convergence but struggles with dynamic adaptation. Therefore, designing a dynamic important weight identification that integrates current and historical importance is paramount for improving training accuracy while retaining robust model convergence. 

In light of this, we define the weight importance of trainable weight $\bf{W}$ at the $t$-th training round, denoted as ${{\bar I}_t}\left( {\bf{W}} \right)$, as a weighted average of historical and current importance:
\begin{equation}\label{eq:weighted_importance}
{{\bar I}_t}\left( {\bf{W}} \right) = {\gamma _t}{{\bar I}_{t - 1}}\left( {\bf{W}} \right) + \left( {1 - {\gamma _t}} \right){{\tilde I}_t}\left( {\bf{W}} \right),
\end{equation}
where $\gamma _t$ is a balance parameter controlling the trade-off between historical and current importance, ${{\bar I}_{t - 1}}\left( {\bf{W}} \right)$ is the cumulative historical importance up to the $(t-1)$-th round, and ${{\tilde I}_t}\left( {\bf{W}} \right) = \Theta\big( {{{\bf{W}}{_{t - 1}}}} \big)$ is the current importance in the $t$-th training round.

The key to identifying important weights lies in the design of the balance parameter $\gamma _t$. To gain deeper insights into its impact on training performance, we fine-tune a single weight with the highest weight importance under varying balance parameters on LLaMA-2-7B. Fig.~\ref{fig:impor_weight_para_final} shows that different balance parameters exhibit distinct convergence speed and training accuracy, while Fig.~\ref{fig:impor_weight_para_cross_round} reveals that the optimal balance parameter varies across training rounds. These observations underscore the necessity of dynamically adjusting balance parameters for each training round.

Motivated by these insights, we formulate the balance parameter $\gamma _t$ by considering two pivotal factors:

\begin{figure}[t]
  \centering
  \subfloat[Converged accuracy and time vs. balance parameter\label{fig:impor_weight_para_final}]{
    \includegraphics[width=0.535\columnwidth]{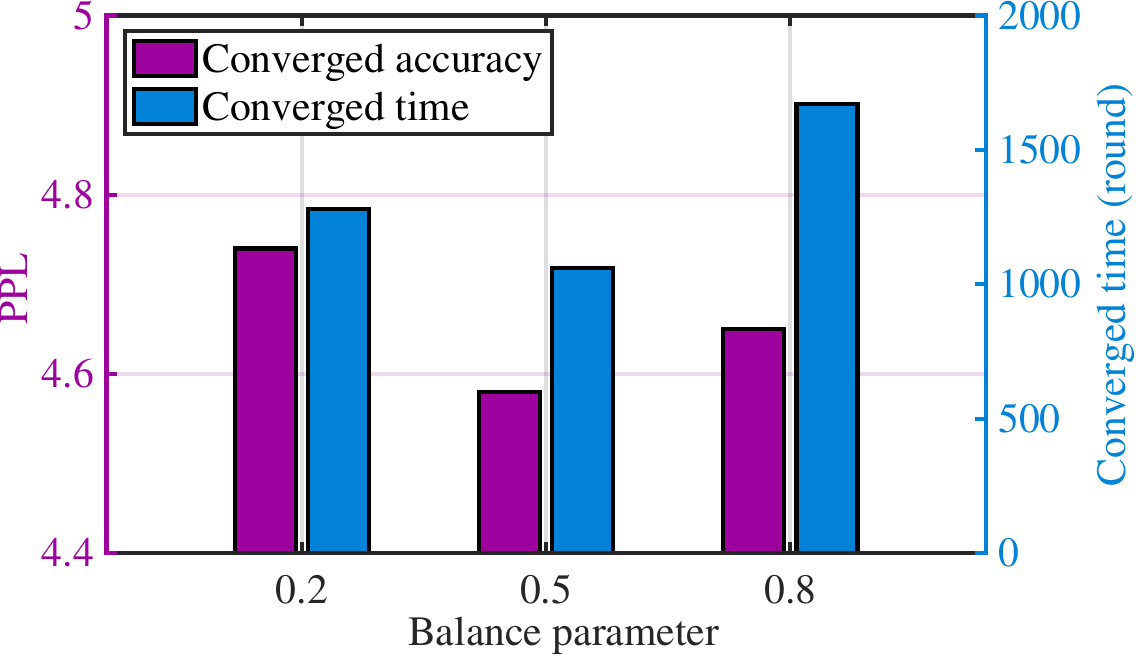}}
  \subfloat[Converged accuracy vs. training round\label{fig:impor_weight_para_cross_round}]
  {
    \includegraphics[width=0.470\columnwidth]{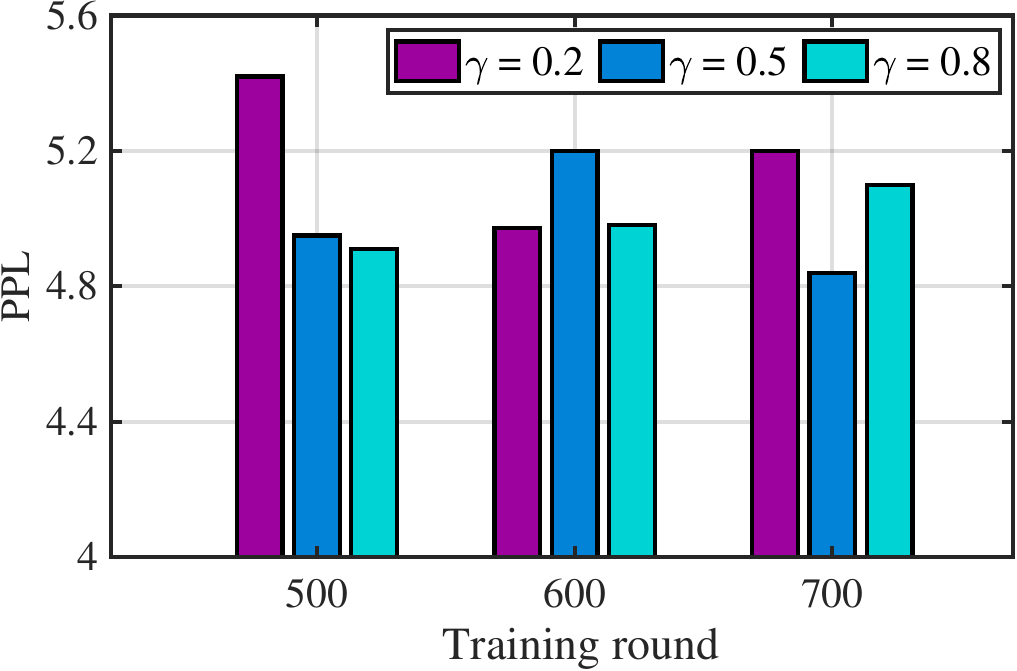}  
  }
  \caption{The converged accuracy and time versus
balance parameter (a) and converged accuracy across 200 training rounds with varying balance parameters (b) on LLaMA-2-7B.
}
  \label{fig:important_balance_para}
\end{figure}

\begin{itemize}
    \item {\bf Training Stage:} In early training stages, the model relies heavily on current importance to adapt to the data due to its limited task understanding. As the model training converges, excessive dependence on current importance may lead to instability or overfitting. The model must gradually shift its focus from current to historical importance. This transition is governed by the ratio of the current round index $t$ to the total training rounds $T$, i.e., $\frac{t}{T}$.
    \item {\bf Weight Importance Fluctuations:} The extreme fluctuations in weight importance can negatively impact training efficiency. When a weight becomes excessively important, it may dominate learning and overfit to data-specific patterns, thereby degrading the model’s generalization ability. Conversely, if a weight’s importance becomes too low, the model may suppress potential information, leading to underfitting or suboptimal convergence. Therefore, it is essential to regularize the weight importance by jointly considering current importance and historical importance. We utilize the importance ratio $\frac{{{{\tilde I}_t \left( {\bf{W}} \right)}}}{{{{\bar I}_{t - 1} \left( {\bf{W}} \right)}}}$ to enable bidirectional adjustment: when the current weight importance exceeds the historical importance, we attenuate its influence to avoid overreacting to transient fluctuations; when it falls below, we amplify its effect to prevent the suppression of useful information, ensuring that weights with low current importance are still given a chance to contribute, especially in early training stages where weight importance may be unreliable.
    \end{itemize}

Based on these two factors, we define the balance parameter $\gamma _t$ for the $t$-th training round as:
\begin{equation}
{\gamma _t} = {{1 - \exp ( - \frac{{{{\tilde I}_t \left( {\bf{W}} \right)}t}}{{{{\bar I}_{t - 1} \left( {\bf{W}} \right)}T}})}}.
\end{equation}

\begin{figure}[t]
  \centering
    \subfloat[Normalized important metric\label{fig:important_ppl}]
  {
    \includegraphics[width=0.472\columnwidth]{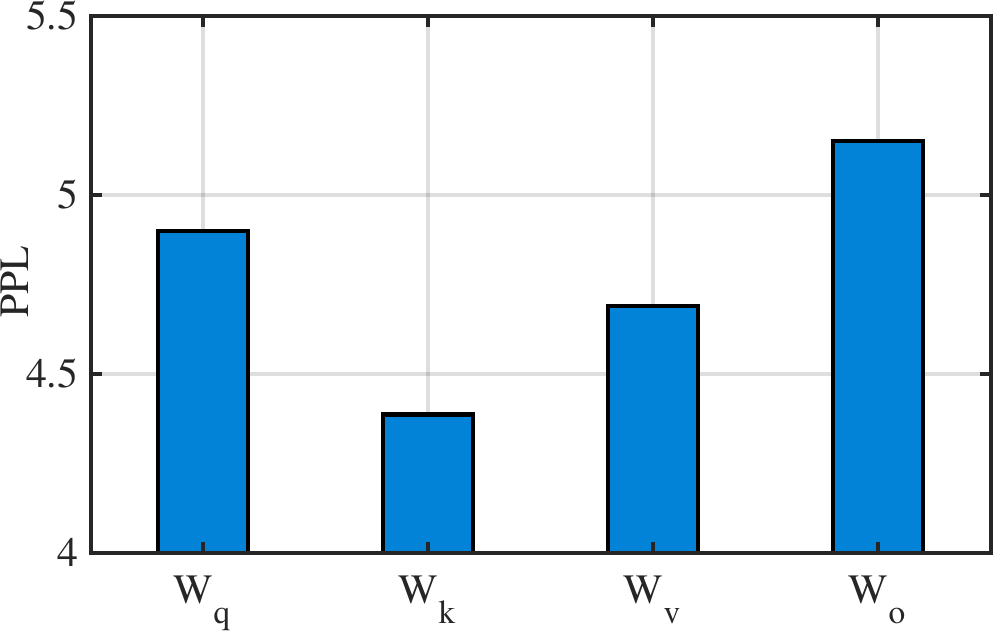}
  }
  \subfloat[PPL\label{fig:important_indicator}]{
    \includegraphics[width=0.474\columnwidth]{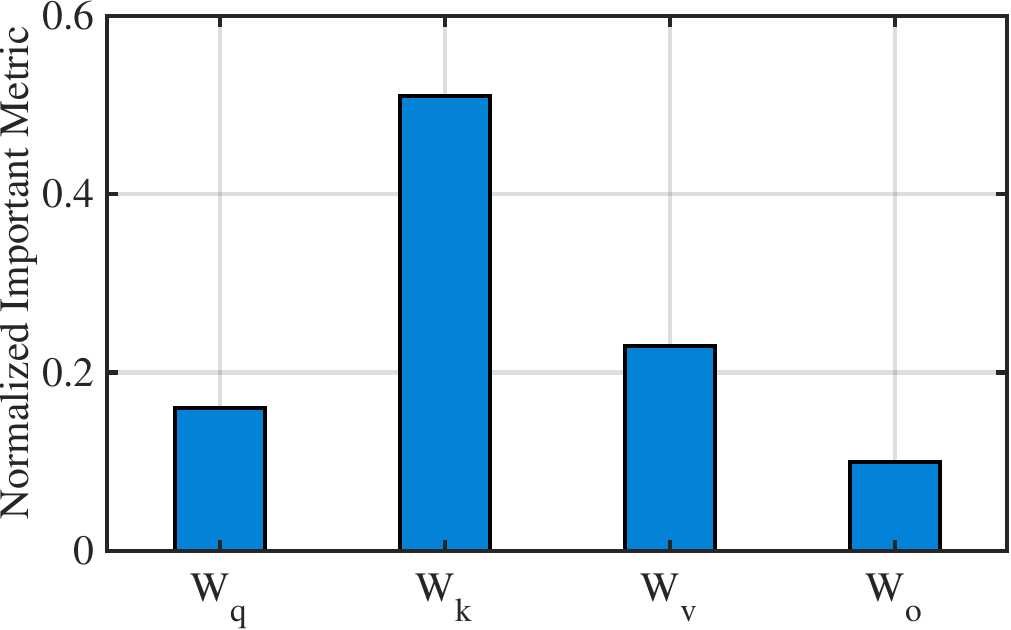}
  }
  \caption{The normalized weight importance and PPL of different trainable weights on LLaMA-2-7B across 200 training rounds.}
  \label{fig:important_metric}
\end{figure}

As shown in Fig.~\ref{fig:important_metric}, we observe that trainable weights with higher weight importance, as defined in~\eqref{eq:weighted_importance}, exhibit better fine-tuning performance, i.e., lower PPL values, validating the effectiveness of the proposed weight importance metric. With the weight importance, client devices and the central server can configure the combinations of LoRA adapters and model splitting, as discussed in the following section.

\subsection{Adaptive Rank and Model Splitting Configuration} \label{ssec:dynamic_split_rank}

Recalling Section~\ref{ssec:bg:straggler}, the computing heterogeneity across client devices causes the severe device unavailability problem, significantly degrading the training performance of LLM SL. In LLM SL, the fine-tuning computing cost on client devices and the central server is jointly determined by the decomposition rank of the LoRA adapters and the model split point. On the one hand, a shallower model split point reduces the computing burden on client devices but shifts substantial load to the central server~\cite{lin2024adaptsfl,lin2024splitlora}. On the other hand, the decomposition rank of the LoRA adapter presents a trade-off between computational cost and fine-tuning performance: A higher rank improves data fitting but incurs more computing resources~\cite{hu2021lora,lin2024splitlora}. The tight coupling of these parameters necessitates a unified optimization solution to enhance fine-tuning performance. 

\begin{figure}[t]
    \centering
    \includegraphics[width=0.9\linewidth]{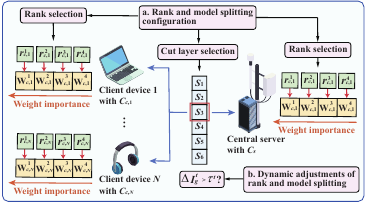}
    \caption{The illustration of adaptive rank and model splitting configuration solution.}
    \label{Optimal_Solution_fed}
     \vspace{-2ex}
\end{figure}

We propose an adaptive rank and model splitting configuration based on the weight importance indicator in Section~\ref{ssec:import_identify}, as shown in Fig.~\ref{Optimal_Solution_fed}. The strategy consists of two key components: i) rank and model splitting configuration, and ii) adaptive adjustments of rank and model splitting.

{\bf a) Rank and Model Splitting Configuration.} Given the vast search space, an exhaustive search over all possible combinations of LoRA adapter ranks and model split points is computationally prohibitive. To circumvent this impasse, we leverage the weight importance indicator developed in Section~\ref{ssec:import_identify} to configure the LoRA adapters and model splitting. For a given model split point $S$, the client devices and the central server first sort trainable weights (shown in Fig.~\ref{fig:eg_fine_tune}) by importance and then configure the LoRA adapters accordingly. Specifically, the LoRA adapters are assigned to trainable weights in descending order of importance, selecting ranks from a predefined set $\mathcal{Q}$. If the computing cost of the selected rank is within the available budget, the rank is selected; otherwise, the next most important weight is evaluated. This process continues until the remaining computing budget is exhausted, yielding an optimal LoRA configuration for the given split point $S$.

To identify the most effective combination of model split point and LoRA adapter configuration, we compute the global weight importance for each candidate combination, which captures the overall importance of weights across both the client-side and server-side LLMs, computed as the sum of the average of the weight importance for selected client-side and server-side weights. The global weight importance for the given model split point $S$ is given by
{ \begin{equation}\label{eq:global_weight_import}
\begin{aligned}
I_g\left( S \right) & = \frac{1}{{\sum\limits_{n = 1}^N \left| \mathcal{W}_{c,n} \right|}} \sum\limits_{n = 1}^N \sum\limits_{\substack{r \in \mathcal{R}_{c,n}, \\ \mathbf{W} \in \mathcal{W}_{c,n}}} \Theta\left( \mathbf{W}_{c,n} \middle| S, r, \mathbf{W} \right) \\
&  + \frac{1}{{\left| \mathcal{W}_s \right|}} \sum\limits_{\substack{r \in \mathcal{R}_s, \\ \mathbf{W} \in \mathcal{W}_s}} \Theta\left( \mathbf{W}_s \middle| S, r, \mathbf{W} \right).
\end{aligned}
\end{equation}}

By comparing the global weight importance across the predefined set of model split points $\mathcal{S}$, the model split point and LoRA adapter with the highest global weight importance are selected. This ensures that the most important weights are prioritized, enhancing training performance while accommodating the heterogeneous computing budgets of the client devices.

\begin{figure}[t]
  \centering
  \small
  \subfloat[PPL\label{fig:model_split_accu}]{
    \includegraphics[width=0.474\columnwidth]{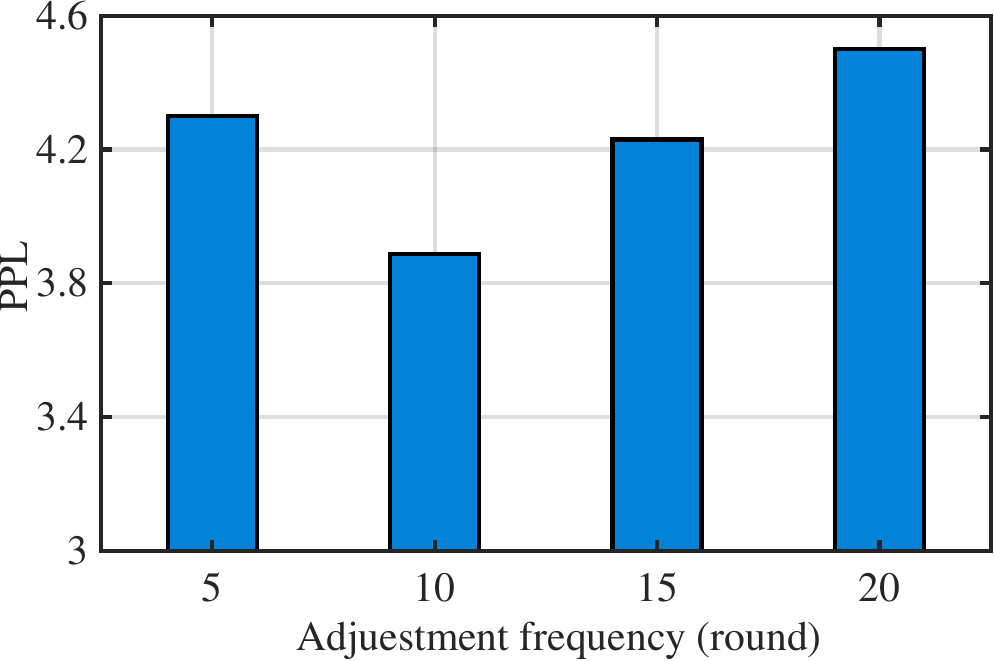}
  }
  \subfloat[STD of PPL\label{fig:model_split_std}]
  {
    \includegraphics[width=0.474\columnwidth]{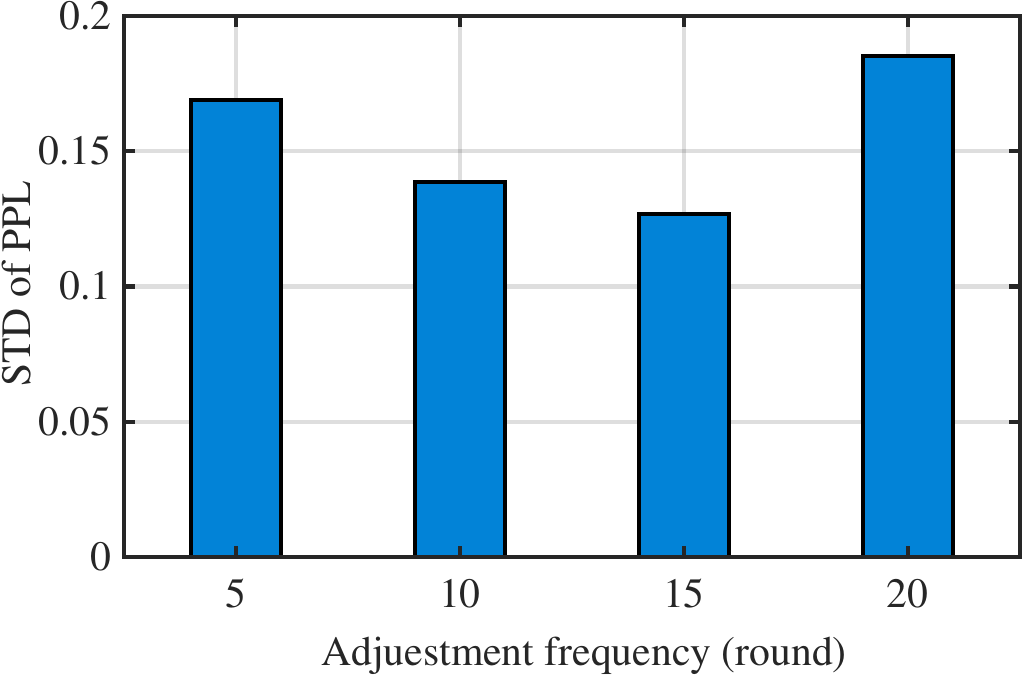}
  }
  \caption{ The PPL and its STDs of fine-tuning LLaMA-2-7B with varying adjustment frequencies.}
  \label{fig:model_split_sensitive}
   \vspace{-1ex}
\end{figure}

{\bf b) Adaptive Adjustments of Rank and Model Splitting.}
As mentioned in Section~\ref{ssec:bg:straggler}, the computing budgets of client devices and the central server vary during model training. To accommodate these fluctuations, dynamic rank and model split point adjustments are essential. Since LoRA fine-tuning updates only a small portion of the whole model parameters\footnote{The fine-tuning parameters are substantially fewer than those of the global model~\cite{hu2021lora,dettmers2023qlora}.}, ranks can be flexibly adjusted in each training round to accommodate the fluctuations in computing budgets without compromising model stability~\cite{hu2021lora}. In contrast, the adjustment frequency of model split point must be carefully determined. Frequent adjustments may degrade fine-tuning performance and stability due to the rapid changes in the overall model aggregation frequency~\cite{lin2024adaptsfl}, whereas infrequent adjustments may hinder the model from adapting to the weight importance variations, leading to a performance decline.

To better understand the impact of adjustment frequency of the model split point on training performance, we fine-tune the LLaMA-2-7B with varying adjustment frequencies. We use the number of training rounds between two consecutive model split point adjustments to represent the adjustment frequency. Fig.~\ref{fig:model_split_accu} demonstrates that a moderate adjustment frequency maximizes accuracy, while Fig.~\ref{fig:model_split_std} shows that both excessively frequent and infrequent adjustments can result in increased STD, thereby degrading training stability.

We propose an adaptive strategy for adjusting the model splitting point based on the maximum global weight importance difference  ${\Delta I}_{g, t}\left( {S} \right)$, as given by:
{ \begin{equation}
\Delta I_{g, t}\left( S \right) = \mathop {\max }\limits_{S' \in \mathcal{S}\backslash \left\{ S \right\}}  I_{g, t}\left( S' \right) - I_{g, t}\left( {S} \right).
\end{equation}}
The model split point adjustment is triggered when ${\Delta I}_{g, t}\left( {S} \right)$ exceeds a predefined threshold $\tau_t$, indicating that the current model splitting configuration exhibits significantly worse performance than alternative options. For substantial importance difference, $\tau_t$ should be set lower to trigger the more frequent model split point adjustment for responding promptly to the importance difference. For minimal importance difference, $\tau_t$ should be appropriately increased to avoid excessive adjustments, ensuring training stability and efficiency. Thus, the threshold is designed as follows:
{ \begin{equation}
{\tau _t} = {\tau _{t - 1}}\max\left( {1 - \Delta {I_{g, t}}\left( S \right), \varepsilon } \right), 
\end{equation}}
where $\varepsilon>0$ is a small positive constant.

Furthermore, when rank adjustments cannot accommodate changes in the computing budgets of the client devices or central server, a model split point adjustment is triggered. This adaptive strategy dynamically accommodates variations in computing resources and weight importance, ensuring the efficiency and stability of the training process. We summarize the procedure of adaptive rank and model splitting configuration in {\bf Algorithm~\ref{alg:adaptive_config}}.

\RestyleAlgo{ruled}
\LinesNumbered
\begin{algorithm} [t]
\caption{Adaptive Rank and Model Splitting Configuration}
\label{alg:adaptive_config}
\setstretch{1.0}
\small
\SetKwInOut{Input}{Require}
\SetKwProg{Fns}{Rank_model_split_config}{:}{}
\SetKwFunction{Fns}{Rank_model_split_config}
\SetKwProg{Fn}{}{:}{}
\SetKwFunction{Fa}{Adaptive_config}
\Input{$\mathcal{N}$, $C_{c,n}$, $C_{s}$, $\mathcal{S}$, $\mathcal{R}_{c, n}$, $\mathcal{R}_{s}$, $\mathcal{W}_{c, n}$, $\mathcal{W}_{s}$. }
\KwData{$\mathcal{D}=\{\mathcal{D}_{1}, \mathcal{D}_{2}, \cdots, \mathcal{D}_{N}\}$ where $\mathcal{D}_{n}$ is the local dataset of the $n$-th client devices.}

\Fn{\Fa{$\mathcal{N}$, $\mathcal{S}$}}{
\For{$t=1,2,...,T$}{
    Compute global weight importance change $\Delta I_g(S_t)$\\
    \If{$\Delta I_g(S_t) > \tau_t$}{
        \For{$n \in \mathcal{N}$ in parallel}{
    {\texttt{Rank_model_split_config(}$C_{c,n,t}$, $\mathcal{R}_{c,n}$, $\mathcal{W}_{c,n}$\texttt{)}}
    }
    {\texttt{Rank_model_split_config(}$C_{s,t}$, $\mathcal{R}_{s}$, $\mathcal{W}_{s}$\texttt{)}}
    }
    \Else{Configure adapter ranks to fit $C_{c,n,t}$ and $C_{s,t}$}
}}

\Fn{\Fns{$C$, $\mathcal{R}$, $\mathcal{W}$}}{
        \For{$S \in \mathcal{S}$} {
                \For{$ {\bf W} \in \mathcal{W}$ (sorted by importance)}{
                    \For{$r \in \mathcal{R}$ (descending order)}{
                        \If{$f({\bf W}, r) \leq C$}{
                            Configure rank $r$ to weight $\bf W$
                            $C \leftarrow C - f ({\bf W}, r)$
                        }
                    }
                }
            
            Compute $I_g(S)$ via Eqn.~\eqref{eq:global_weight_import}
        }
        
        Select optimal split point ${S^*} = \arg {\max _{S \in \mathcal{S}}}{I_g}(S)$
    }

\end{algorithm}

\subsection{Noise-free Adapter Aggregation and Deployment} \label{ssec:aggregate}

Current SL frameworks update the client-side model via a weighted average of client-side model parameters~\cite{thapa2022splitfed,lin2024efficient}. However, for LLMs with billions of parameters, transmitting client-side LLM for model aggregation is bandwidth-intensive and overwhelms the limited computing capabilities of client devices~\cite{lin2024splitlora} (see Section~\ref{ssec:bg:peft}). 
LoRA fine-tuning mitigates this issue by encoding model updates as low-rank decomposition matrices, allowing only these matrices to be transmitted for model aggregation. However, aggregating low-rank matrices remains challenging. As discussed in Section~\ref{ssec:bg:heter_adapter_agg}, the heterogeneous computing budgets across client devices result in low-rank matrices with varying ranks, rendering direct numerical averaging infeasible. Furthermore, the state-of-the-art aggregation scheme~\cite{zhang2024towards} is mathematically inconsistent with the aggregation of client-side LLM
updates, introducing additional noise into the adapter aggregation process.

\begin{figure}[t]
\centering
\includegraphics[width=0.85\linewidth]{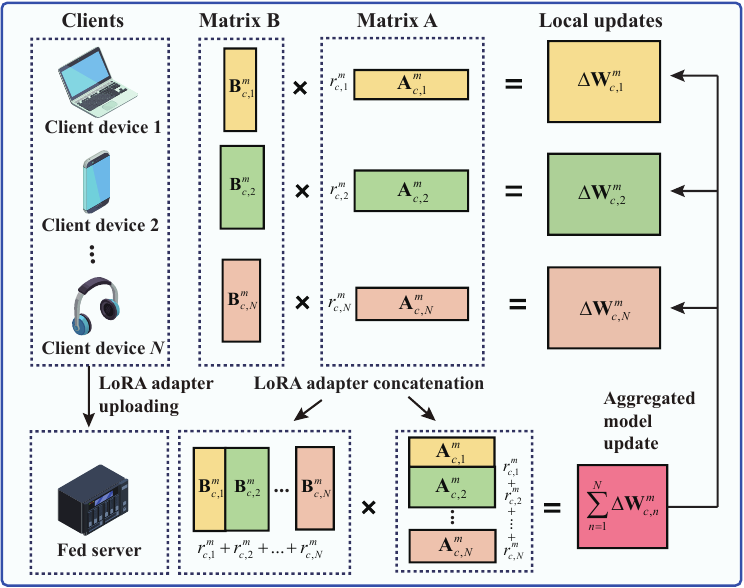}
\caption{The noise-free adapter aggregation scheme of \name framework, where ${\bf{B}}_{c,n}^m$ and ${\bf{A}}_{c,n}^m$ represent low-rank decomposition matrices of the $m$-th trainable weight for client device $n$, and its corresponding decomposition rank is denoted by ${r}_{c,n}^m$.
}
\label{fig:aggregation}
\end{figure}

To combat these challenges, as illustrated in Fig.~\ref{fig:aggregation}, we propose a noise-free adapter aggregation scheme that employs matrix concatenation to enable seamless aggregation of heterogeneous low-rank matrices without introducing additional noise. The proposed scheme consists of the following two stages.

{\bf a) LoRA Adapter Concatenation.} Though the LoRA adapters received by the fed server exhibit structural heterogeneity, we observe that the total ranks of low-rank decomposition matrices $\bf B$ and $\bf A$ across client devices remain consistent. This insight suggests that structural consistency can be achieved by concatenating multiple LoRA adapters along the rank dimensions of $\bf B$ and $\bf A$.

As illustrated in Fig.~\ref{fig:aggregation}, at the $t$-th training round, the concatenated decomposition matrices of the $m$-th client-side trainable weight (LoRA adapter) are expressed as 
{ \begin{equation}
{{\bf{B}}_c^{m,t} \!= \![{\bf{B}}_{c,1}^{m,t},{\bf{B}}_{c,2}^{m,t}, \cdots ,{\bf{B}}_{c,N}^{m,t}] \!\in \! {\mathbb{R}^{{d_i} \times (r_{c,1}^{m,t} + r_{c,2}^{m,t} + \!...\! + r_{c,N}^{m,t})}}}
\end{equation}}
and 
{ \begin{equation}
{{\bf{A}}_c^{m,t} = \left[ \!\!{\begin{array}{*{20}{c}}
{{\bf{A}}_{c,1}^{m,t},}\\
{{\bf{A}}_{c,2}^{m,t},}\\
 \vdots \\
{{\bf{A}}_{c,N}^{m,t}}
\end{array}} \!\!\right] \in {\mathbb{R}^{(r_{c,1}^{m,t} + r_{c,2}^{m,t} + ... + r_{c,N}^{m,t}) \times {d_o}}},}
\end{equation}}
where $d_i$ and $d_o$ denote the input and output dimensions of the low-rank decomposition matrices $\bf B$ and $\bf A$, respectively. Therefore, at the $t$-th training round, the product of the concatenated decomposition
matrices, representing the aggregated incremental model update, can be calculated as~\cite{cho2024heterogeneous}
{ \begin{equation}\label{intre_update}
\Delta {\bf{W}}_{c}^{m,t} 
\!=\! [{\bf{B}}_{c,1}^{m,t},{\bf{B}}_{c,2}^{m,t}, \cdots ,{\bf{B}}_{c,N}^{m,t}]\!\!\left[\!\! {\begin{array}{*{20}{c}}
{{\bf{A}}_{c,1}^{m,t},}\\
{{\bf{A}}_{c,2}^{m,t},}\\
 \vdots \\
{{\bf{A}}_{c,N}^{m,t}}
\end{array}} \!\!\right] \!\!= \!\!\sum\limits_{n = 1}^N {{\bf{B}}_{c,n}^{m,t}{\bf{A}}_{c,n}^{m,t}}.  
\end{equation}}

Recall the incremental model update $\Delta {\bf{W}}_{c,n}^m = {{\bf{B}}_{c,n}^m{\bf{A}}_{c,n}^m}$ from Section~\ref{ssec:bg:heter_adapter_agg}, the product of the concatenated decomposition matrices is equivalent to the sum of the client-side incremental model updates from the client devices. This indicates the noise-free property of the concatenate-then-multiply scheme, ensuring accurate LoRA aggregation without introducing additional noise.

{\bf b) Incremental Matrix Update.} While the concatenate-then-multiply scheme enables the computation of the aggregated incremental model update via Eqn.~\eqref{intre_update}, recovering the low-rank decomposition matrices for each client device is non-trivial. The matrix decomposition is usually non-unique and often yields multiple valid solutions. Moreover, the aggregated incremental model update contains mixed information across LoRA adapters, this mixing effect obscures the unique characteristics of individual adapters, rendering accurate reconstruction of updated adapters for each client device difficult~\cite{hyeon2021fedpara}. Therefore, instead of directly updating the decomposition matrices ${\bf{B}}_{c,n}^m$ and ${\bf{A}}_{c,n}^m$, we merge the aggregated incremental model update $\Delta {\bf{W}}_{c}^m$ into the pre-trained client-side model ${\bf{W}}_{c}^m$, as given by
{ \begin{equation}\label{eq:merge_update}
{\bf{W}}_c^{m,t + 1} = {\bf{W}}_c^{m,t} + \Delta {\bf{W}}_c^{m,t}.
\end{equation}}
After that, ${\bf{B}}_{c,n}^{m, t}$ is initialized to zero and ${\bf{A}}_{c,n}^{m, t}$ is initialized~\cite{hu2021lora,singhal2024exact} using a random Gaussian distribution at the beginning of the next training round. 

\section{Implementation} \label{sec:impl}
In this section, we first elaborate on the implementation of \name and then introduce the experiment setup.

\subsection{Implementing \name}

\begin{figure}[t]
\centering
\includegraphics[width=0.95\linewidth]{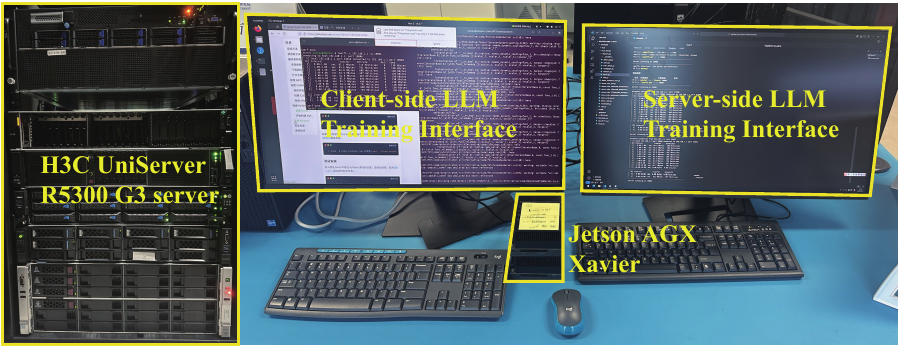}
\caption{\name prototype and testbed.}
\label{fig:testbed_H}
\end{figure}

We prototype \name for LLM fine-tuning based on a micro services architecture, as illustrated in Fig.~\ref{fig:testbed_H}. The platform consists of Jetson development kits and a high-performance server, with network conditions controlled via \textit{tc}~\cite{beshay2015fidelity}. The central server is emulated by an H3C UniServer R5300 G3 server equipped with eight NVIDIA GeForce RTX 3090 GPUs, dual Intel Xeon Silver 4210R processors (10 cores, 2.84 GHz each), and 8×32 GB DDR4 RAM, running Ubuntu 18.04.6 LTS. The Jetson series kits~\cite{nvidia_jetson2024} have been widely recognized as a standard hardware platform for edge AI applications, offering on-device computing resources (e.g., GPU acceleration for deep learning inference and training) within a constrained power budget. Thus, to emulate the client devices, we utilize the Jetson AGX Xavier kits, equipped with a 512-core Volta GPU with Tensor Cores, an 8-core ARM V8.2 64-bit CPU, 32 GB EMMC 5.1 storage, and also running Ubuntu 18.04.6 LTS. The software stack comprises Python 3.7 and PyTorch 1.9.1, which are used for implementing LLM fine-tuning for natural language generation applications.

\subsection{Experiment Setup}

{\bf{Datasets and Models.}} We adopt the  E2E dataset~\cite{novikova2017e2e} to evaluate the performance of \name. The E2E dataset focuses on the restaurant domain, consisting of 42,000 training, 4,600 validation, and 4,600 test samples. To implement \name, we employ two well-known LLMs, LLaMA-2-7B~\cite{touvron2023llama} and GPT-2-L~\cite{radford2019language}. The LLaMA-2-7B model comprises 32 layers and 3.52 billion parameters, while the GPT-2-L features 24 layers with 355 million parameters and 36 layers with 774 million parameters. In our experiments, we fine-tune the LLaMA-2-7B and GPT-2-L models on the E2E dataset for the natural language generation (NLG) task.

{\bf{Benchmarks.}} To comprehensively evaluate the performance of \name, we compare \name against the following alternatives:
\begin{itemize}
    \item {\bf Full-parameter fine-tuning (FT):} Central server fine-tunes parameters of the whole LLM, allowing the model to fully adapt to new tasks or datasets while leveraging its pre-trained knowledge~\cite{kenton2019bert}. 
    \item {\bf Centralized low-rank adaptation (CenLoRA):} Central server updates LoRA adapters of the whole LLM with the full dataset~\cite{hu2021lora}.
    \item {\bf Heterogeneous federated low-rank adaptation (HetLoRA):} HetLoRA allows client devices to fine-tune LLM with varying LoRA rank via FL and employ a sparsity-weighted aggregation scheme to effectively aggregate LoRA adapters~\cite{cho2024heterogeneous}.
    \item {\bf Split low-rank adaptation (SplitLoRA):} SplitLoRA partitions the LLM into client-side and server-side LLMs, deployed on the client devices and central server for fine-tuning, and periodically aggregates client-side LoRA adapters~\cite{lin2024splitlora}.     
\end{itemize}

\begin{figure}[t]
  \centering
  \subfloat[LLaMA-2-7B (homo). \label{fig:convergence_homo_llama}]
  {
    \includegraphics[width=0.469\columnwidth]{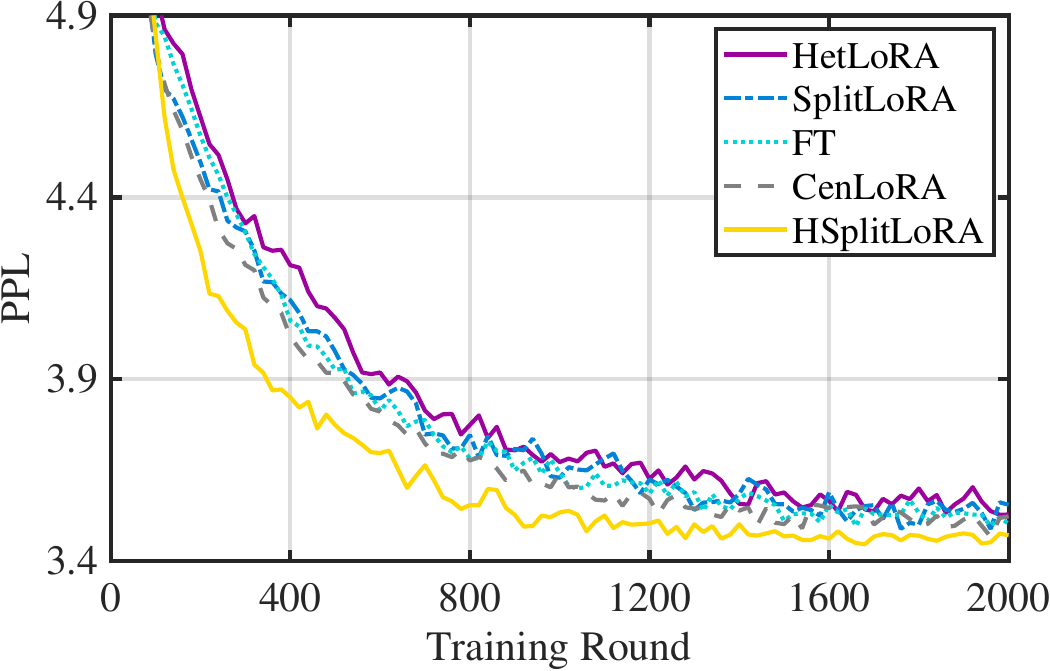}
  }
    \subfloat[LLaMA-2-7B (hetero). \label{fig:convergence_hetero_llama}]
  {
    \includegraphics[width=0.469\columnwidth]{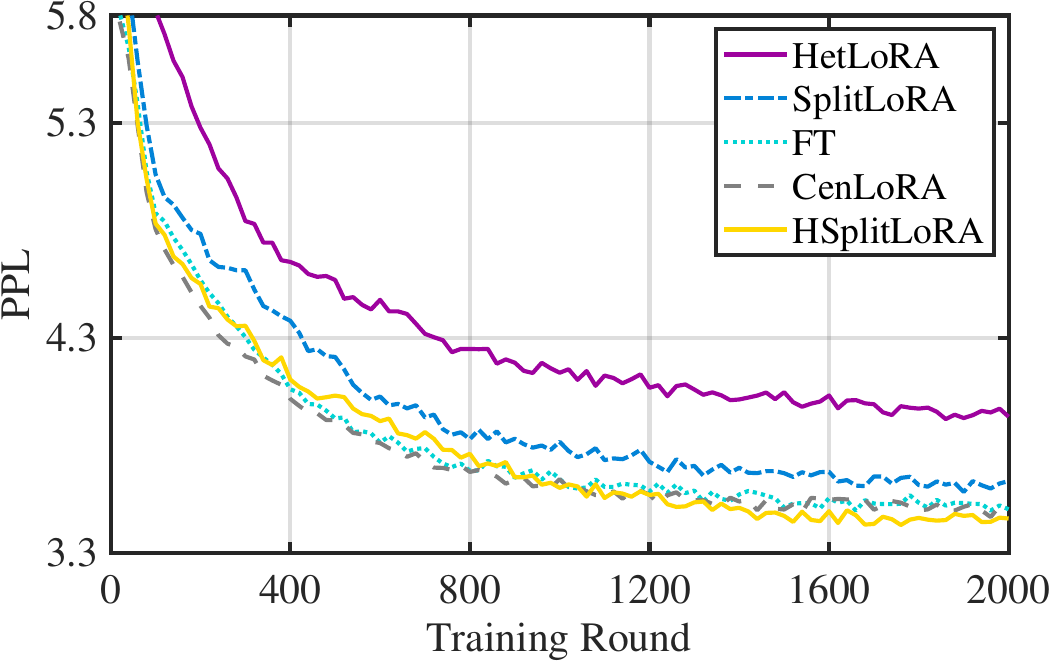}
  } \\
  \subfloat[GPT-2-L (homo).
  \label{fig:convergence_homo_gpt2}]
  {
    \includegraphics[width=0.469\columnwidth]{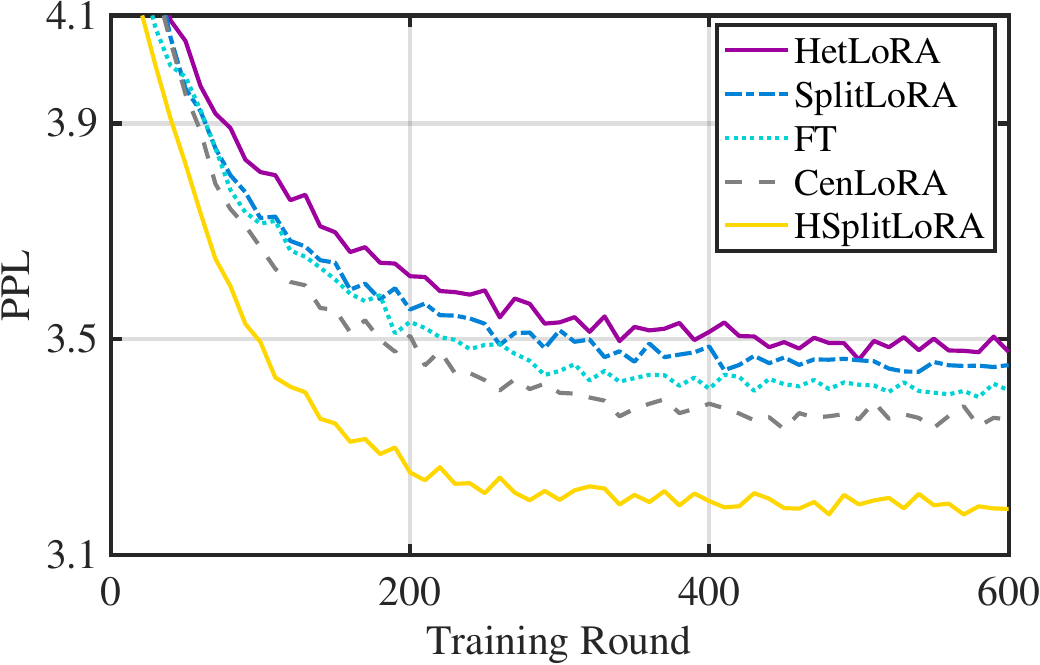}
  }
  \subfloat[GPT-2-L (hetero). \label{fig:convergence_hetero_gpt2}]
  {
    \includegraphics[width=0.469\columnwidth]{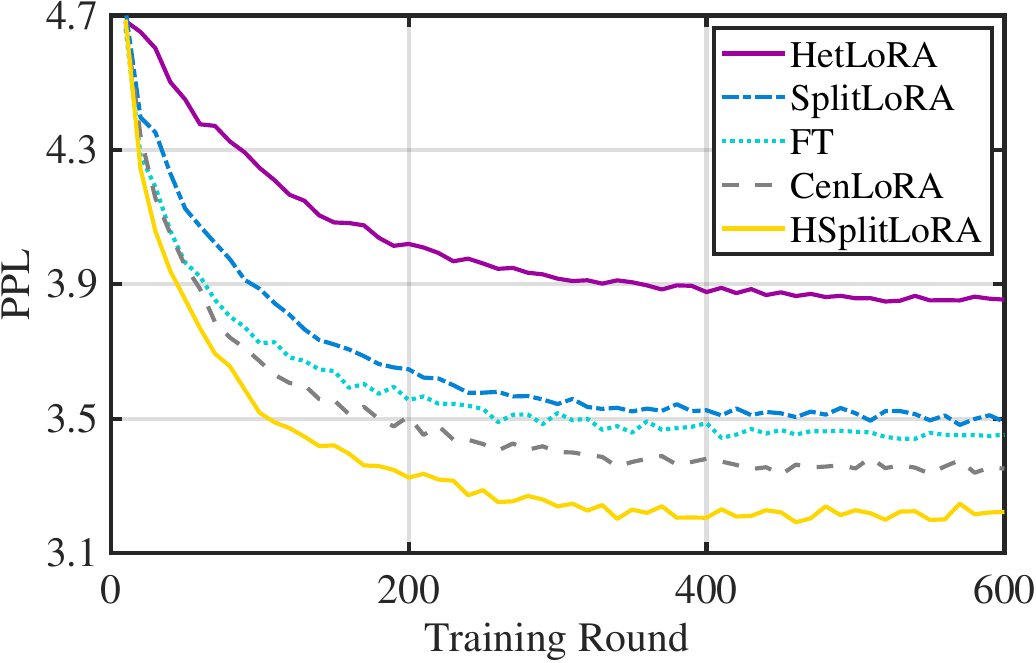}
  }
  \caption{The training performance on LLaMA-2-7B and GPT-2-L models under heterogeneous (hetero) and homogeneous (homo) settings.}
  \label{fig:overall_performance}
\end{figure}

{\bf{Hyper-parameters.}} For the NLG task on the E2E dataset for LLaMA-2-7B, we set the mini-batch size, learning rate, and maximum sequence length to 1, 0.0001, and 384, respectively, while these parameters are configured as 4, 0.0002, and 512 for GPT-2-L. The total number of participating client devices is 5, all running in a synchronous mode~\cite{ho2013more}. For the homogeneous setting, we set the maximum GPU memory of each client device to 20GB for LLaMA-2-7B and 4GB for GPT-2-L. For the heterogeneous setting, the maximum GPU memory of client devices follows a uniform distribution between 5GB and 20GB for LLaMA-2-7B, and between 1GB and 4GB for GPT-2-L. For both homogeneous and heterogeneous settings, the maximum GPU memory of the central server is set to 75GB for LLaMA-2-7B and 15GB for GPT-2-L. The predefined rank set is $\mathcal{Q}=\{1, 2, 4, 8, 16, 32\}$.

\section{Performance Evaluation} \label{sec:eval}
This section provides numerical results to evaluate the training performance of \name framework and the effectiveness of each meticulously designed component.

\subsection{Superiority of \name}

\begin{figure}[t]
  \centering
  \subfloat[LLaMA-2-7B \label{fig:convergence_statics_llama}]{
    \includegraphics[width=0.495\linewidth]{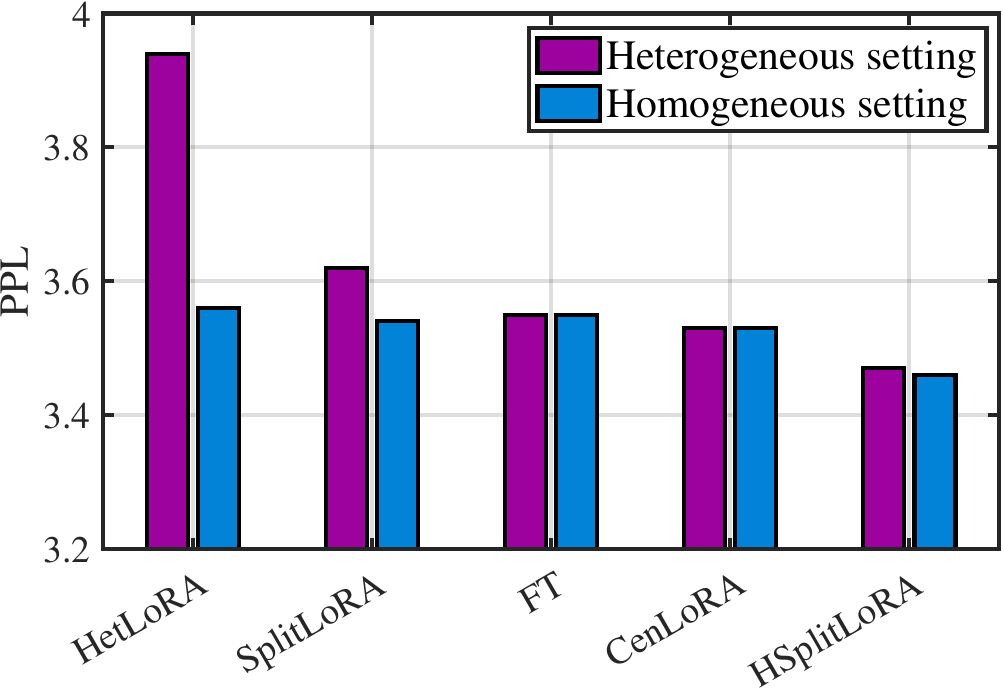}
  }
  \subfloat[GPT-2-L\label{fig:convergence_statics_gpt2}]
  {
    \includegraphics[width=0.492\linewidth]{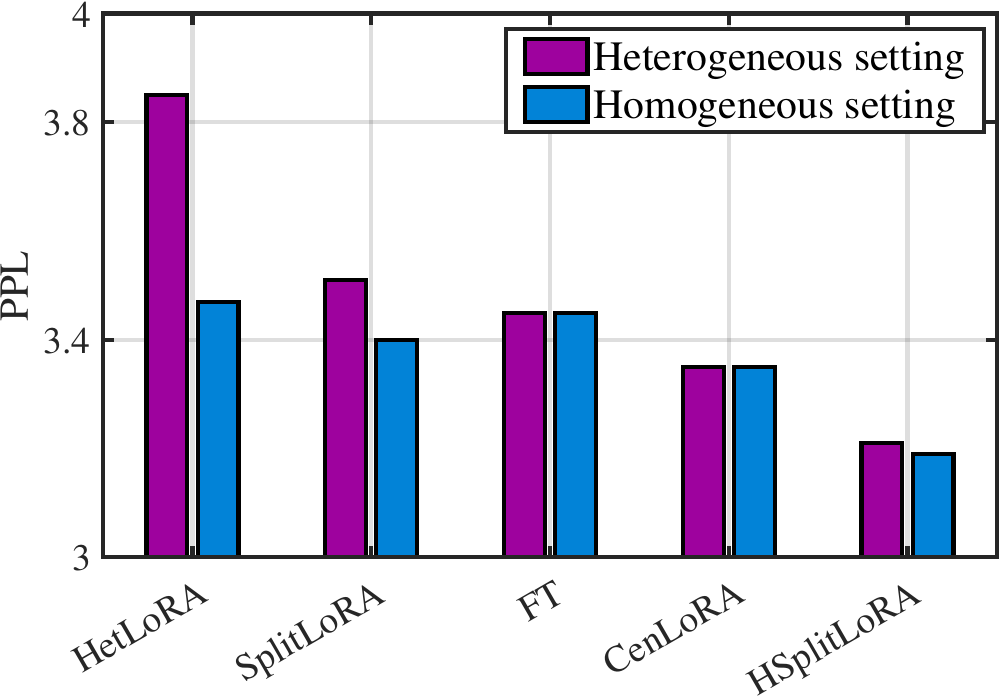}
  }
  \caption{The converged accuracy on LLaMA-2-7B and GPT-2-L models under heterogeneous and homogeneous settings.}
  \label{fig:converged_acc}
   \vspace{-2ex}
\end{figure}

\begin{figure}[t]
  \centering
  \subfloat[LLaMA-2-7B \label{fig:convergence_statics_llama}]{
    \includegraphics[width=0.495\linewidth]{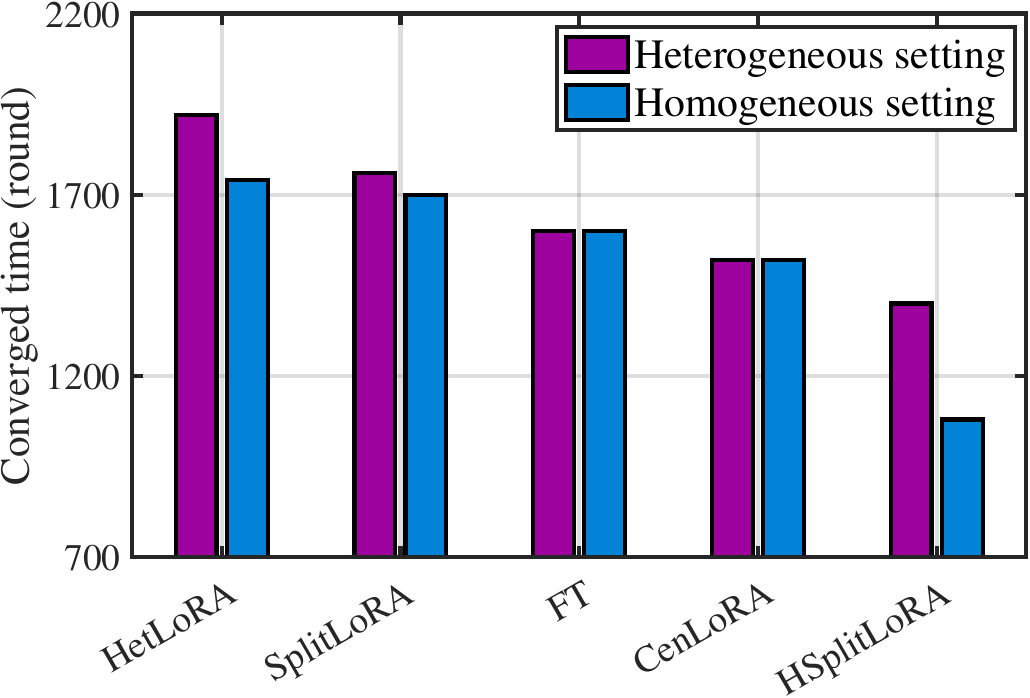}
  }
  \subfloat[GPT-2-L\label{fig:convergence_statics_gpt2}]
  {
    \includegraphics[width=0.495\linewidth]{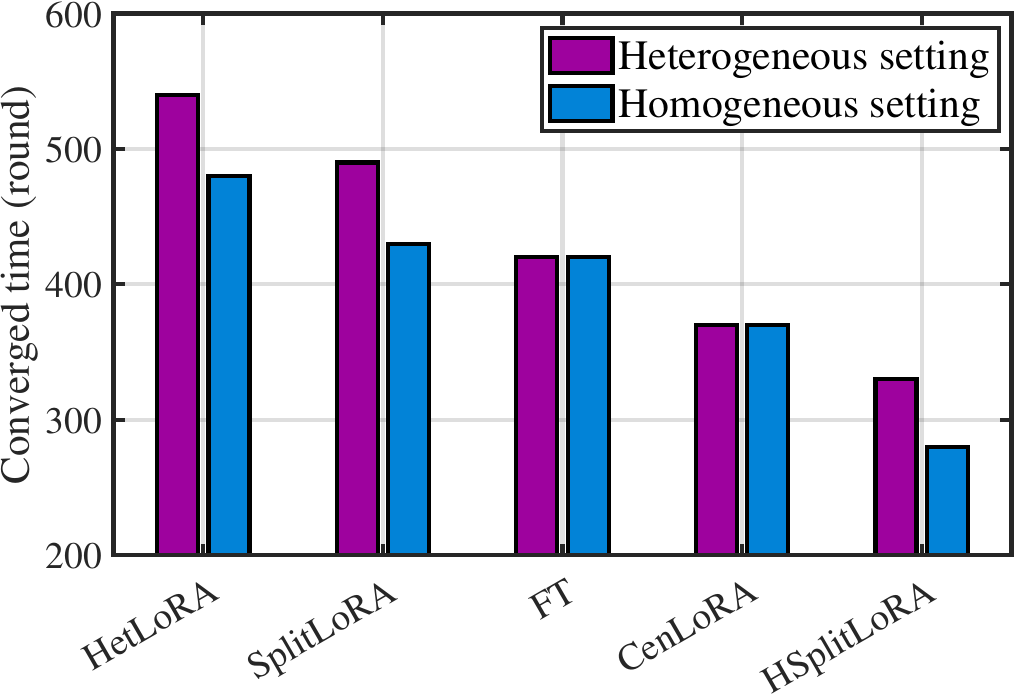}
  }
  \caption{The converged time on LLaMA-2-7B and GPT-2-L models under heterogeneous and homogeneous settings.}
  \label{fig:converged_time}
\end{figure}

\begin{table*}[t]
\centering
\setlength{\tabcolsep}{12pt} %
\renewcommand{\arraystretch}{0.87} %
\resizebox{0.83\textwidth}{!}{ %
\begin{tabular}{c c c c c c c}
\toprule
\multirow{2}{*}{\textbf{Model}} & \multirow{2}{*}{\textbf{Method}} & \multicolumn{5}{c}{\textbf{E2E NLG Challenge}} \\
\cmidrule(lr){3-7} 
& & \textbf{BLEU} & \textbf{NIST} & \textbf{MET} & \textbf{ROUGE-L} & \textbf{CIDEr} \\
\midrule
\multirow{5}{*}{LLaMA-2-7B (homo)} 
 & FT       &  66.4 &  8.45 &  44.6&  69.0& 2.35 \\
 & CenLoRA    & 66.7 &  8.46 & 44.8 & 69.1 & 2.37 \\
 & HetLoRA      &66.2	&8.42	&44.3	&68.8	&2.34 \\
 & SplitLoRA       &  66.5&  8.44& 44.7	 & 69.0 &  2.36\\
  & \bf{\name (Ours)}  & \bf{68.1} & \bf{8.66} & \bf{45.7} & \bf{69.8} & \bf{2.45} \\
\midrule
\multirow{4}{*}{LLaMA-2-7B (hetero)}
 & FT       &  66.4 &  8.45 &  44.6&  69.0& 2.35 \\
 & CenLoRA    & 66.7 &  8.46 & 44.8 & 69.1 & 2.37 \\
 & HetLoRA       &61.3 &8.07 &41.1 &64.7 &2.09  \\
  & SplitLoRA       &65.4  & 8.38 & 44.0 &  68.4& 2.32 \\
 & \bf{\name (Ours)}      & \bf{68.0} & \bf{8.65} & \bf{45.6} & \bf{69.5} & \bf{2.42} \\
\midrule
\multirow{5}{*}{GPT-2-L (homo)}
 & FT       &  68.2&  8.68&  45.8&  69.9&  2.45\\
 & CenLoRA    &68.9&  8.76&  46.5&  70.7&  2.47  \\
 & HetLoRA     &68.0  &8.64  &45.6 & 69.5 & 2.42 \\
  & SplitLoRA   &68.6  &8.79  &46.3 & 70.2 & 2.46  \\
 & \bf{\name (Ours)}     & \bf{69.7} & \bf{8.82} & \bf{46.8} & \bf{71.2} & \bf{2.51} \\
\midrule
\multirow{5}{*}{GPT-2-L  (hetero)}
 & FT       &  68.2&  8.68&  45.8&  69.9& 2.45 \\
 & CenLoRA      &68.9&  8.76&  46.5&  70.7&  2.47  \\
 & HetLoRA       &62.2 &8.17 &41.5 &65.2 &2.12  \\
  & {SplitLoRA}   &67.2 &8.57 &45.4 &69.4 &2.42\\
 & \bf{\name (Ours)}     & \bf{69.5} & \bf{8.79} & \bf{46.6} & \bf{70.9} & \bf{2.49} \\
\bottomrule
\end{tabular}
}
\caption{The performance comparison of various metrics on LLaMA-2-7B and GPT-2-L models  under heterogeneous and homogeneous settings.}
 \vspace{-1ex}
\end{table*}

{\bf{Training Performance of \name}.} Fig.~\ref{fig:overall_performance} presents the superior training performance of \name compared to four benchmarks under heterogeneous and homogeneous settings. In the homogeneous setting, \name achieves the fastest convergence and highest training accuracy, outperforming FT, CenLoRA, HetLoRA, and SplitLoRA. This advantage stems from \name's important weight identification, which identifies and selectively fine-tunes trainable parameters based on their contributions to LLM training, thereby enhancing training performance. In the heterogeneous setting, computing discrepancies across client devices exacerbate the device unavailability effect in HetLoRA and SplitLoRA, leading to degraded converged accuracy. While HetLoRA mitigates the device unavailability effect by assigning varying ranks of LoRA adapters, its fine-tuning remains inefficient due to the necessity of fine-tuning the whole LLM on the client device. SplitLoRA employs a static rank and model splitting configuration, causing a mismatch between the computing workload and the computing budget of the client device, significantly slowing down model convergence and leading to worse converged accuracy than FT and CenLoRA. In contrast, \name dynamically configures the optimal LoRA adapter ranks and model splitting as model training progresses and computing budgets change, ensuring superior training performance even under the heterogeneous setting. For GPT-2-L, \name still retains a consistent advantage over all benchmarks, demonstrating superior scalability across diverse LLM architectures. These results underscore \name's superior adaptability to heterogeneous edge computing environments, improving convergence speed and accuracy.

{\bf{Converged Accuracy of \name}.} 
Fig.~\ref{fig:converged_acc} illustrates the converged accuracy on LLaMA-2-7B and GPT-2-L under homogeneous and heterogeneous settings. In the heterogeneous setting, HetLoRA and SplitLoRA exhibit a significant performance drop, with PPL increasing by approximately 0.38 and 0.1 for LLaMA-2-7B, and 0.38 and 0.11 for GPT-2-L, compared to the homogeneous setting. The performance degradation of HetLoRA stems from its static LoRA adapter configuration, which lacks dynamic rank adjustments and weight prioritization, leading to inefficiencies under varying computational constraints. SplitLoRA suffers from accuracy loss due to its reliance on fixed LoRA adapters and model splitting, making it susceptible to device unavailability effects and reducing adaptability in heterogeneous settings. In contrast, \name achieves the lowest PPL and outperforms the FT, CenLoRA, HetLoRA, and SplitLoRA for PPL by 0.09 (resp. 0.26), 0.07 (resp. 0.16), 0.1 (resp. 0.28), and 0.08 (resp. 0.21) in the homogeneous setting, and 0.08 (resp. 0.24), 0.06 (resp. 0.14), 0.47 (resp. 0.64), and 0.15 (resp. 0.33) in the heterogeneous setting on LLaMA-2-7B (resp. GPT-2-L). This superiority is attributed to the design of important weight identification to prioritize the fine-tuning of key weights, and the adaptive rank and model splitting configuration to accommodate heterogeneous computing constraints.

Furthermore, \name benefits from noise-free adapter aggregation, facilitating efficient aggregation of client-side updates from client devices without introducing additional noise. While \name, HetLoRA, and SplitLoRA showcase noticeable accuracy degradation in heterogeneous settings, FT and CenLoRA remain unaffected as their high computing demands confine execution to the central server. However, their reliance on centralized training severely limits their  scalability, rendering them impractical for large-scale distributed learning. We also conduct a comprehensive comparison of the accuracy across various metrics in homogeneous and heterogeneous settings, with the results summarized in Table I.

\begin{figure}[t]
  \centering
  \subfloat[LLaMA-2-7B\label{fig:trainable_parameters_llama}]{
    \includegraphics[width=0.497\linewidth]{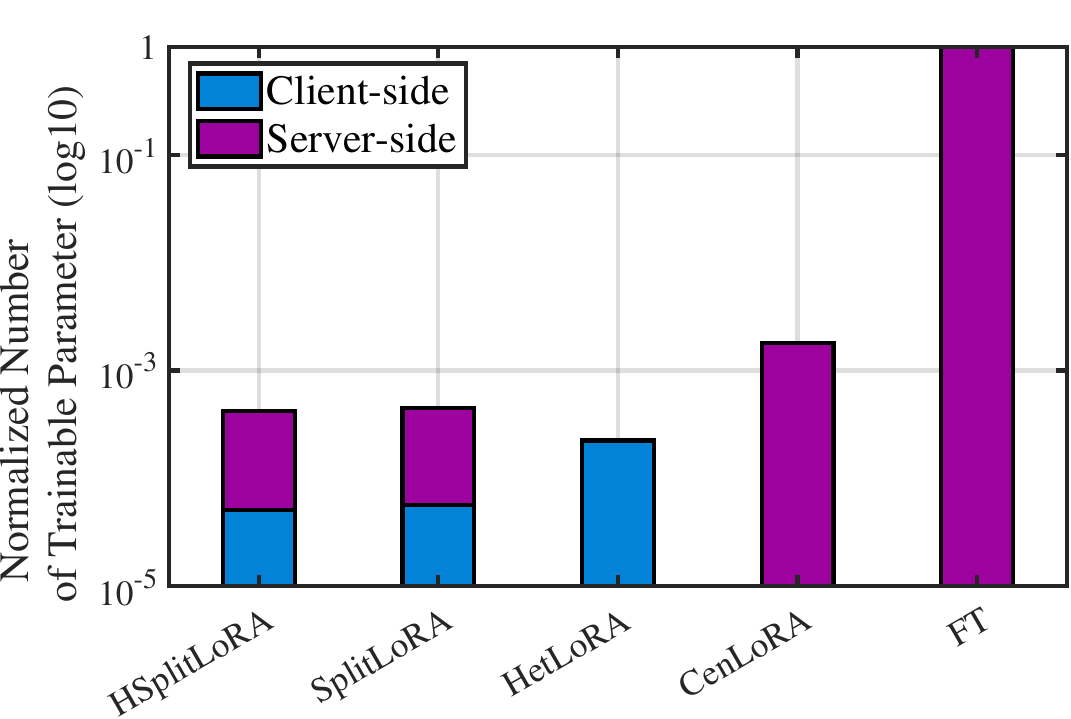}
  }
  \subfloat[GPT-2-L\label{fig:trainable_parameters_gpt2}]
  {
    \includegraphics[width=0.495\linewidth]{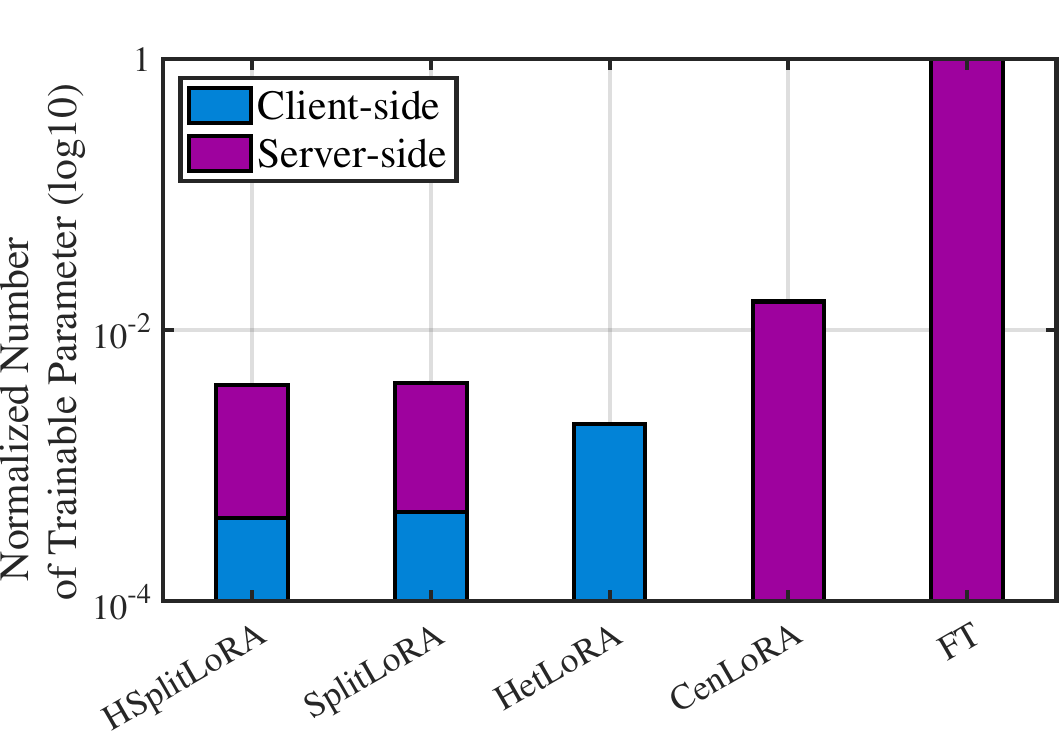}
  }
  \caption{The number of trainable parameters for LLaMA-2-7B and GPT-2-L models. }
  \label{fig:trainable_parameters}
   \vspace{-2ex}
\end{figure}

\begin{figure*}[t]
  \centering
  \subfloat[IWI\label{fig:hetero_weights}]{
    \includegraphics[width=0.29\linewidth]{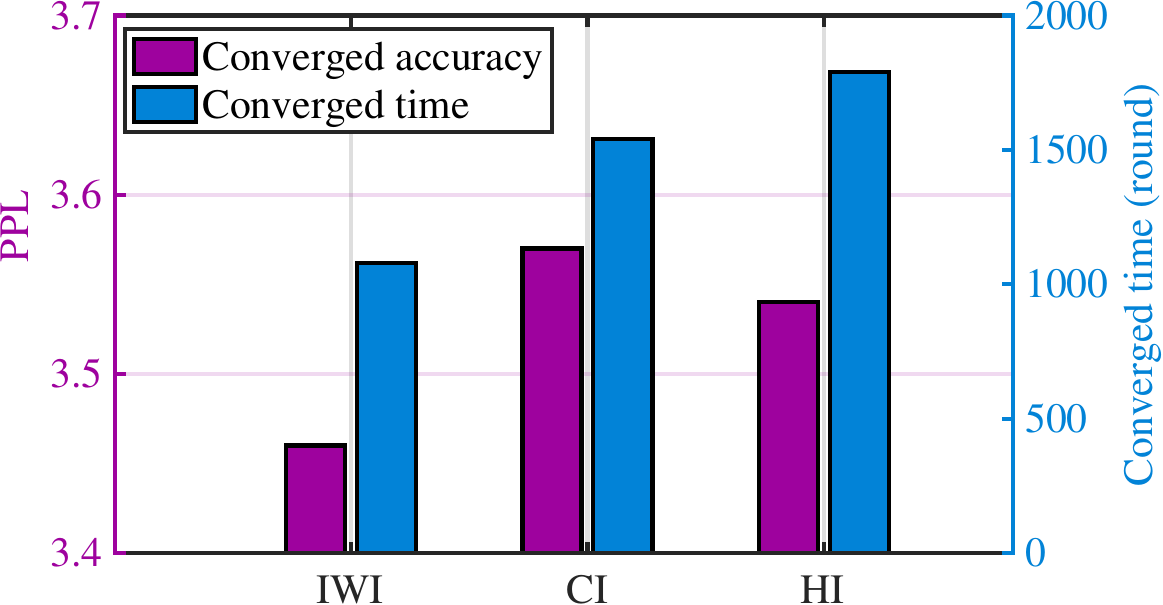}
  }
  \subfloat[ARMSC\label{fig:hetero_avg_var_gpt2}]
  {
    \includegraphics[width=0.29\linewidth]{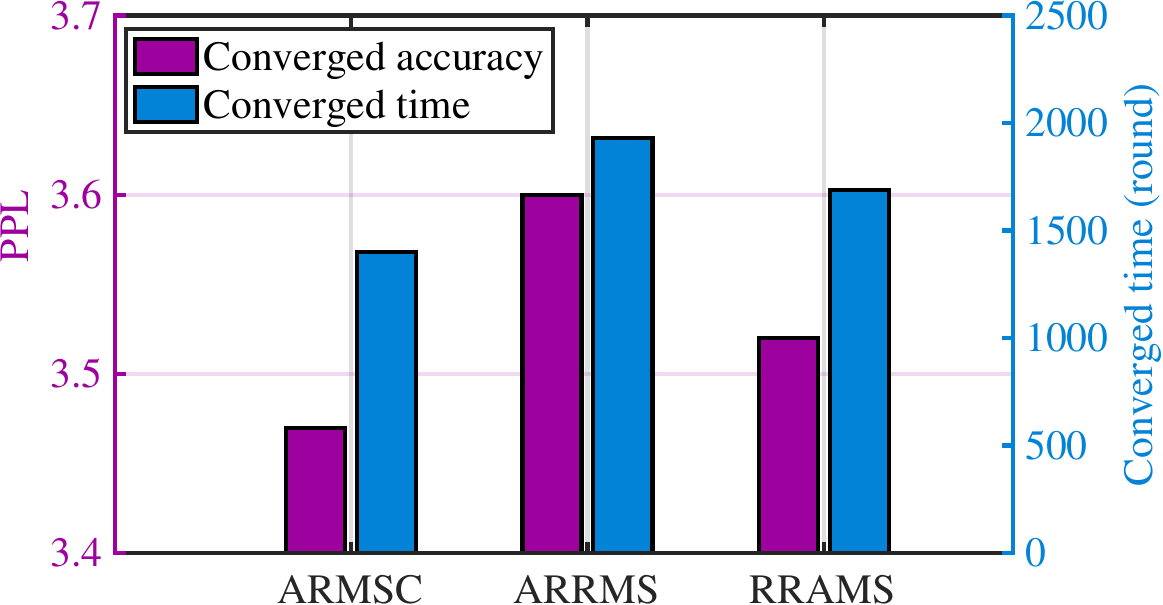}
  }
    \subfloat[NAA\label{fig:adapter_free_aggre}]
  {
    \includegraphics[width=0.29\linewidth]{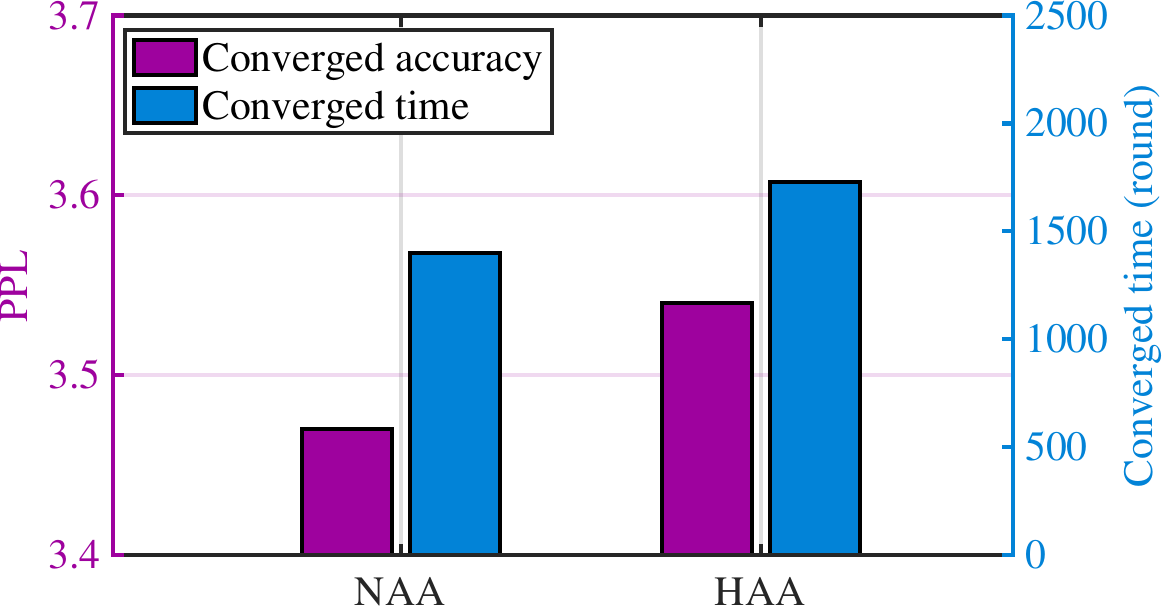}
  }
  \caption{The ablation evaluation for IWI (a), ARMSC (b), and NAA (c) on LLaMA-2-7B.}
  \label{fig:ppl_gpt2}
\end{figure*}

{\bf{Converged Time of \name}.} 
Fig.~\ref{fig:converged_time} compares the convergence time of \name against four benchmarks on LLaMA-2-7B and GPT-2-L under the heterogeneous and homogeneous settings. In the homogeneous setting, \name consistently exhibits the fastest convergence speed, surpassing FT, CenLoRA, HetLoRA, and SplitLoRA by factors of approximately 1.4, 1.3, 1.6 and 1.5 for LLaMA-2-7B, and by factors of 1.5, 1.3, 1.7, and 1.5 for GPT-2-L, respectively. This performance gain stems from our design of the important weight identification strategy, which prioritizes the fine-tuning of key weights to expedite model training. FT exhibits the slowest convergence due to the full-parameter fine-tuning, whereas CenLoRA achieves slightly faster model convergence by shrinking the number of trainable parameters via LoRA fine-tuning. 
 In the heterogeneous setting, HetLoRA and SplitLoRA suffer from severe device unavailability effects caused by computing discrepancies across client devices, resulting in slower convergence speed. \name still maintains the shortest converged time in the heterogeneous setting. This is primarily attributed to adaptive rank and model splitting configuration, which dynamically accommodates heterogeneous computing budgets across client devices.

\begin{table}[t]
\centering
\small
\renewcommand{\arraystretch}{0.87} 
\resizebox{0.5\textwidth}{!}{ 
\begin{tabular}{l c c c c c}
\toprule
\multirow{2}{*}{\textbf{Method}} & \multicolumn{5}{c}{\textbf{E2E NLG Challenge}} \\
\cmidrule(lr){2-6} 
 & \textbf{BLEU} & \textbf{NIST} & \textbf{MET} & \textbf{ROUGE-L} & \textbf{CIDEr} \\
\midrule
CI   &66.1	&8.40	&44.2	&68.7	&2.32 \\
HI   & 66.6 & 8.44 & 44.7 & 69.1 & 2.37 \\
\textbf{IWI (Ours)}                & \bf{ 68.1} & \bf{8.66} & \bf{45.7} & \bf{69.8} & \bf{2.45} \\
\bottomrule
\end{tabular}
}
\caption{The performance comparison of IWI and counterparts
relying solely on current importance (CI) or historical importance (HI) on LLaMA-2-7B.}
\label{tab:weight_index}
\end{table}

\begin{table}[t]
\centering
\small
\renewcommand{\arraystretch}{0.87} 
\resizebox{0.5\textwidth}{!}{ 
\begin{tabular}{l c c c c c}
\toprule
\multirow{2}{*}{\textbf{Method}} & \multicolumn{5}{c}{\textbf{E2E NLG Challenge}} \\
\cmidrule(lr){2-6} 
 & \textbf{BLEU} & \textbf{NIST} & \textbf{MET} & \textbf{ROUGE-L} & \textbf{CIDEr} \\
\midrule
ARRMS    & 65.6 & 8.39 & 44.2 & 68.5 & 2.33 \\
RRAMS    & 66.8&  8.47 & 44.8&  69.2&  2.38 \\
\textbf{ARMSC (Ours)}                & \bf{68.0} & \bf{8.65} & \bf{45.6} & \bf{69.5} & \bf{2.42} \\
\bottomrule
\end{tabular}
}
\caption{The performance comparison of ARMSC, random rank and adaptive model splitting (RRAMS), and adaptive rank and random model splitting (ARRMS) on LLaMA-2-7B.}
\label{tab:hc}
\end{table}

{\bf{Number of Trainable Parameters}.} Fig.~\ref{fig:trainable_parameters} presents the average number of trainable parameters of \name and four benchmarks methods for fine-tuning LLaMA-2-7B and GPT-2-L. It is shown that FT exhibits the highest number of trainable parameters, exceeding CenLoRA by at least 550 times on LLaMA-2-7B and by approximately 60 times on GPT-2-L. This is because FT involves updating all parameters of LLM, whereas CenLoRA only updates the parameters of LoRA adapters, which constitute only a small fraction of the global model. HetLoRA has fewer trainable parameters than CenLoRA, as client devices have limited computing resources and cannot support the fine-tuning of the same number of trainable parameters as the central server. \name and SplitLoRA exhibit substantially fewer client-side trainable parameters than HetLoRA. This reduction stems from model splitting, which offloads most trainable parameters to the central server and reduces the computing workload on client devices. Moreover, since \name and SplitLoRA only transmit the client-side LoRA adapters to the fed server for aggregation, a small number of trainable parameters indicates a lower volume of data transmissions for adapter aggregation, thus enhancing fine-tuning efficiency.

\subsection{Ablation Study}

{\bf{Important Weights Identification (IWI).}} Fig.~\ref{fig:hetero_weights} compares the training performance of IWI with counterparts relying solely on current importance (CI) or historical importance (HI) on LLaMA-2-7B under the homogeneous setting. The results demonstrate that IWI achieves the lowest PPL while maintaining a convergence speed comparable to CI-based method. This stems from IWI’s adaptive weight importance evaluation, which dynamically captures weight variations during training and balances current and historical importance. In contrast, CI-based methods adjust rapidly but are vulnerable to noise and local optima, while the HI-based method exhibits long-term stability but responds sluggishly to dynamic weight importance changes. The comparison of accuracy across various metrics is presented in Table~\ref{tab:weight_index}.

\begin{table}[t]
\centering
\small
\renewcommand{\arraystretch}{0.87} 
\resizebox{0.5\textwidth}{!}{ 
\begin{tabular}{l c c c c c}
\toprule
\multirow{2}{*}{\textbf{Method}} & \multicolumn{5}{c}{\textbf{E2E NLG Challenge}} \\
\cmidrule(lr){2-6} 
 & \textbf{BLEU} & \textbf{NIST} & \textbf{MET} & \textbf{ROUGE-L} & \textbf{CIDEr} \\
\midrule
HAA     & 66.6&   8.45 &44.8& 69.1 &2.36 \\
\bf{NAA (Ours)}        & \bf{68.0} & \bf{8.65} & \bf{45.6} & \bf{69.6} & \bf{2.43} \\
\bottomrule
\end{tabular}
}
\caption{The performance comparison of NAA and homogeneous adapter aggregation (HAA)~\cite{zhang2024towards} on LLaMA-2-7B.}
\label{tab:adapter_free_aggre}
\end{table}

{\bf{Adaptive Rank and Model Splitting Configuration (ARMSC).}} Fig.~\ref{fig:hetero_avg_var_gpt2} illustrates the training performance of ARMSC compared to random rank and adaptive model splitting (RRAMS) and adaptive rank and random model splitting (ARRMS) on LLaMA-2-7B under the heterogeneous setting. The results indicate that ARMSC achieves the shortest convergence time and lowest PPL. This is attributed to its ability to adjust the LoRA adapter ranks and model split points to accommodate heterogeneous computing budgets across client devices. By adapting LoRA ranks to available computing resources of client devices, ARMSC prevents both overloading and underutilization, while adaptive model splitting balances the computing load between client devices and the central server to mitigate the device unavailability effect. Furthermore, ARMSC leverages global weight importance change as a triggering indicator, ensuring that model split points are updated only when significant changes in weight importance occur, avoiding frequent unnecessary adjustments that could destabilize training. Table~\ref{tab:hc} provides training performance on other metrics for further comparison.

{\bf{Noise-free Adapter Aggregation (NAA).}} Fig.~\ref{fig:adapter_free_aggre} presents the training performance of NAA and homogeneous adapter aggregation (HAA)~\cite{zhang2024towards} on LLaMA-2-7B under the homogeneous setting. NAA consistently outperforms HAA in both accuracy and convergence speed, achieving a lower PPL and a shorter convergence time. This superiority stems from NAA’s concatenation-based low-rank adapter aggregation, which preserves mathematical consistency and avoids aggregation-induced noise. In contrast, HAA averages low-rank adapters across client devices before matrix multiplication to approximate the aggregated client-side LLM update. This operation introduces additional noise, degrading training performance. A more comprehensive comparison of NAA and HAA across various metrics is illustrated in Table~\ref{tab:adapter_free_aggre}.

\section{Related Work} \label{sec:rw}

{\bf{Parameter-efficient Fine-tuning.}} PEFT has become a promising approach for efficient LLM fine-tuning, substantially reducing trainable parameters without compromising performance. Among various PEFT approaches, Adapter~\cite{houlsby2019parameter, pfeiffer2020adapterfusion, karimi2021compacter} and LoRA~\cite{hu2021lora, sheng2023s} are the most popular two. Adapter introduces lightweight trainable modules in each transformer block, which are fine-tuned while freezing the pre-trained model. LoRA enhances fine-tuning efficiency by decomposing weight updates into two low-rank matrices while keeping the pre-trained model frozen. LoRA has become the dominant PEFT method due to its significant superiority over Adapter, i.e., the substantial reduction of trainable parameters through low-rank decomposition without introducing additional inference latency~\cite{hu2021lora, hu2023llm, sheng2023s}. Several LoRA variants have been proposed to enhance the efficiency and adaptability in LLM fine-tuning~\cite{hayou2024lora,zhou2024lora,kopiczko2023vera}, such as LoRA+~\cite{hayou2024lora} that introduces distinct learning rates for low-rank matrices, and LoRA-drop~\cite{zhou2024lora} that prunes less important adapters based on output importance to reduce trainable parameters. VeRA~\cite{kopiczko2023vera} utilizes random projections to initialize low-rank matrices with shared weights, reducing computing overhead without sacrificing training performance.

{\bf{Split Learning.}} With the growing demand for efficient distributed learning systems, SL~\cite{vepakomma2018split} become a compelling distributed framework but suffers from excessive training latency due to its sequential training from one client device to another. To overcome this severe limitation, SFL~\cite{thapa2022splitfed} and parallel SL~\cite{kim2023bargaining} are proposed to parallelize SL for expediting model training. To further improve the training efficiency of SL, some variants of SL have been developed~\cite{lin2024efficient,wu2023split,pal2021server,lin2024adaptsfl}. EPSL~\cite{lin2024efficient} reduces the dimension of activations’ gradients via last-layer gradient aggregation to accelerate model training, while CPSL~\cite{wu2023split} partitions devices into several clusters to reduce training latency. SGLR~\cite{pal2021server} averages the local gradients at the cut layer to combat the server-side large batch and the backward client decoupling problems. AdaptSFL~\cite{lin2024adaptsfl} adaptively controls
model splitting and client-side model aggregation to balance communication-computing latency and training convergence.

{\bf{Split Learning for LLMs. }} Developing an efficient SL fine-tuning framework for LLMs is still in its infancy. Our prior work, SplitLoRA~\cite{lin2024splitlora}, was the first split parameter-efficient fine-tuning framework for LLMs, built on the SFL framework and LoRA fine-tuning. Another recent framework, SplitLLM~\cite{zhang2025splitllm}, proposes a hierarchical cloud-edge-client SL scheme for fine-tuning LLM. Though SplitLoRA and SplitLLM have made progress in SL fine-tuning, they fail to scale effectively on computing entities with heterogeneous computing budgets and cannot adjust the fine-tuning strategy to accommodate dynamic changes in computing resources during model training.

\section{Conclusion} \label{sec:con}
Taking an important step towards split parameter-efficient fine-tuning paradigm, we have proposed a heterogeneous parameter-efficient fine-tuning framework built on split learning and LoRA fine-tuning method, named \name. \name consists of three primary components: important weight identification, dynamic rank and model splitting configuration, and noise-free adapter aggregation. First, the important weight identification scheme identifies important weights based on their contributions to LLM training to efficiently fine-tune LLM under computing resource constraints. Second, the dynamic rank and model splitting configuration dynamically adjusts the decomposition ranks of LoRA adapters and model split point to accommodate the heterogeneous computing budgets of client devices. Lastly, the noise-free adapter aggregation mechanism meticulously concatenates low-rank decomposition matrices to support heterogeneous adapter aggregation without introducing additional noise. Extensive experiments demonstrate that \name achieves superior performance compared to  state-of-the-art benchmarks.

\ifCLASSOPTIONcaptionsoff
  \newpage
\fi

\bibliographystyle{IEEEtran}
\bibliography{reference}

\end{document}